\documentclass{article}
\usepackage{iclr2027_conference}
\usepackage{times}
\usepackage[T1]{fontenc}
\usepackage[utf8]{inputenc}
\usepackage{amsmath,amssymb,mathtools}
\usepackage{graphicx}
\usepackage{booktabs}
\usepackage{multirow}
\usepackage{microtype}
\usepackage{xcolor}
\usepackage[colorlinks=true,linkcolor=blue,citecolor=blue,urlcolor=blue]{hyperref}

\newcommand{\I}{I}
\newcommand{\G}{G}
\newcommand{\Rint}{R}
\newcommand{\Dint}{D}
\newcommand{\erasure}{cross-term erasure}
\newcommand{\csafety}{code{+}safety}
\newcommand{\cmath}{code{+}math}
\newcommand{\headline}{task pairs whose local cross-term generation differs by at most 1.9$\times$ differ by 14$\times$--337$\times$ in causally removable interference}

\title{Reading the Gate, Not the Interference: Output-Side Interference Measurement Does Not Track Merge Collapse}

\iclrfinalcopy
\author{Chencheng Zhu \\ UNSW Sydney \\ \texttt{chencheng.zhu@student.unsw.edu.au}}

\makeatletter
\renewcommand{\maketitle}{%
  \par\begingroup
  \newpage\global\@topnum\z@
  {\centering\LARGE\bfseries\MakeUppercase{\@title}\par}%
  \vskip 0.12in
  {\centering \@author \par}%
  \vskip 0.15in
  \endgroup\thispagestyle{plain}}
\makeatother

\begin{document}
\maketitle

\begin{abstract}
Model merging by task arithmetic works until it doesn't, and the field diagnoses why by measuring interference inside the merged model. We take the most direct such measure---the exact layerwise activation cross-term of a factorial ledger---establish its causal anatomy, and then ask what it tracks. The anatomy is clean: each block mostly transports and amplifies the cross-term it receives rather than generating it anew; erased, it is regenerated by the untouched marginal paths to 99\% of its norm unless removed near the output; its output effect varies monotonically with the displacement's angle---orthogonal displacements make interference \emph{worse}---and a two-assumption model derives the angle law, retro-dicts the dose curve ($R^2 = 0.99$--$1.00$), and passes a preregistered scaling test. What the measure tracks is not what the field assumes. Behavioural expert-likeness is decoupled from it at the per-prompt level across four instruments. Its cross-condition behaviour is denominator-dominated: an instruction template pins the main effect to within 1\% while the absolute interaction grows 111$\times$ from two to six merged tasks, so the same wrapper suppresses expressed interference at $k{=}2$ and amplifies it at $k{=}6$. And where merging actually collapses, the cross-term is a \emph{bystander}, not the carrier: across two collapse parameterizations at two scales---even erased \emph{persistently}, at every position---removing it entirely repairs none of the collapse (every preregistered repair gate failed). There the output-side ratio carries no method information under a common counterfactual, while two state-space measures the field already uses rank methods \emph{correctly} at both scales, on GSM8K and HumanEval. All 81 predictions were frozen before their data; falsifications, ours included, are reported as such. Output-side interference measurement, we conclude, reads the gate, the denominator, and the displacement budget---not the interference. What fails a merge is the \emph{carrier--bystander split}: collapse rides in the marginal displacements and the cross-term merely accompanies it---and only state space sees the carrier.
\end{abstract}

\section{Introduction}
\label{sec:intro}

\begin{figure}[b]
\centering
\includegraphics[width=0.92\textwidth]{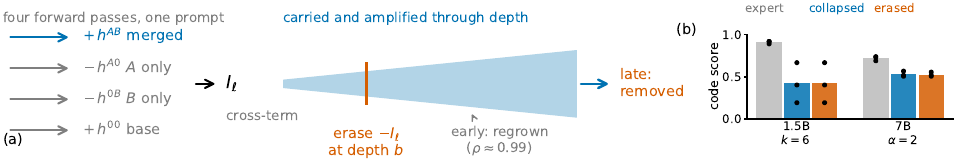}
\vspace{-3mm}
\caption{\textbf{The carrier--bystander split.} \textbf{(a)} Four forward passes over one prompt isolate the exact layerwise cross-term $\I_\ell$; blocks mostly transport and amplify it, so erasure is undone by propagation unless applied near the output, where it is causally effective for output non-additivity (Figure~\ref{fig:one}). \textbf{(b)} Where merging actually collapses it is a \emph{bystander}: erasing all of it leaves the collapsed score unchanged (bars: seed means; dots: seeds) in both parameterizations---collapse rides in the marginal displacements (\S\ref{sec:gating}, \S\ref{sec:metrics}).}
\label{fig:setup}
\vspace{-3mm}
\end{figure}

\begin{figure}[t]
\centering
\includegraphics[width=0.75\textwidth]{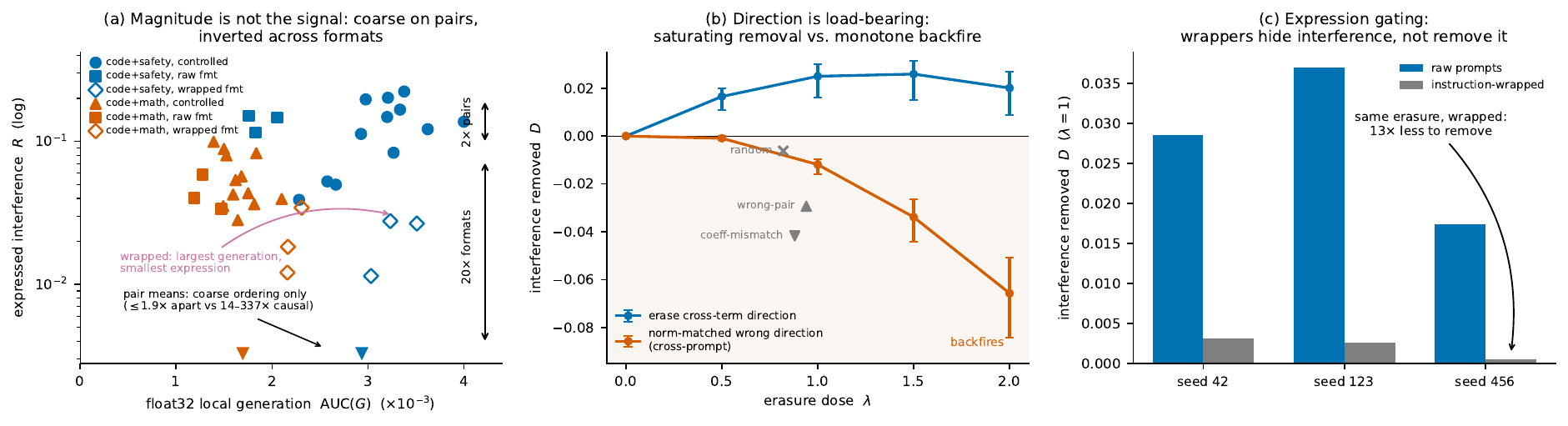}
\caption{\textbf{Orientation, not magnitude: \headline{} (b), and instruction wrappers amplify internal interference while suppressing its expression 13-fold (a, c).}
\textbf{(a)} Float32 local generation AUC($\G$) (from the preregistered fp32 grid, Appendix~\ref{app:decomposition}) versus expressed interference $\Rint$, all 36 cells at the strongest composition point. Magnitude orders the pairs only coarsely (cluster means $\le$1.9$\times$ apart; triangles on the axis) and is \emph{inverted} across formats: wrapped conditions (open diamonds) generate the most and express the least. \textbf{(b)} Dose--response of \erasure{}: removing the cross-term along its own direction ($h^{AB}_\ell - \lambda \I_\ell$) saturates near full erasure; the norm-matched cross-prompt direction backfires monotonically; grey markers are single-dose structural controls. Error bars: seed range. \textbf{(c)} The identical erasure removes 13$\times$ less interference under an Alpaca instruction wrapper---the wrapper hides interference rather than removing it.}
\label{fig:one}
\vspace{-1mm}
\end{figure}

Task arithmetic is disarmingly effective: add two finetuned weight displacements to a base model and, much of the time, both skills transfer \citep{ilharco2023task}. When it fails, the failure is measured at the output---but \emph{located} nowhere. A growing literature quantifies merging interference by the size of internal deviations: layerwise representation bias between merged and expert models \citep{yang2024surgeryv2}, departures from cross-task linearity \citep{zhou2024ctl}, subspace overlap between task matrices \citep{gargiulo2025tsv}. These measurements are proliferating faster than they are audited, and share one untested assumption: \emph{that a merged model measured to carry more interference is a worse merge}. The assumption has consequences: in our collapse cells, the method that best preserves behaviour carries the \emph{highest} measured interference---a practitioner trusting the measure ships the wrong model. What they causally control, whether they track behaviour, and whether they rank merging methods correctly are all open. We settle all three for the most direct such measure.

This paper audits that assumption. We track the exact non-additive residue of merging---the layerwise cross-term $\I_\ell = h^{AB}_\ell - h^{A0}_\ell - h^{0B}_\ell + h^{00}_\ell$---through every block of a merged LLM, measure what each block \emph{generates} versus what it \emph{carries}, erase the carried term surgically to establish what it causally controls, and then ask what that control is worth. The question, precisely:
\begin{center}
\emph{The field measures interference inside merged models. What does that measurement actually track?}
\end{center}

The anatomy comes first (Figure~\ref{fig:one}): the cross-term is mostly transported rather than generated anew, erasing it is undone by regeneration from the untouched marginal paths unless the erasure is applied near the output, and its causal effect on the output follows the displacement's orientation---orthogonal displacements make interference \emph{worse}---with a two-assumption model deriving the law. Then the turn: none of that structure transfers to the quantities the field cares about. The measure is decoupled from behavioural expert-likeness, its cross-condition value is dominated by the denominator a format sets, and where merging collapses it carries no method signal while the family's state-space members rank methods correctly at both scales (\S\ref{sec:metrics}).

Our thesis, in one sentence: \emph{merge collapse rides in the marginal displacements, and the activation cross-term---causally clean, transported, regenerated, erasable---merely accompanies it; measure the carrier, not the bystander.}

\paragraph{Contributions.}
\begin{itemize}\itemsep1pt
\item \textbf{The carrier--bystander split, and what to measure instead (\S\ref{sec:gating}--\ref{sec:metrics}).} Across two collapse parameterizations at two scales, erasing the entire activation cross-term repairs none of the collapse (every repair gate failed; final-boundary behaviour bitwise unchanged in 5/6 cells): collapse rides in the marginal displacements, and only state space sees the carrier. Under a shared reference the output-side ratio carries no method signal (range $<$2\% against behavioural spreads of 0.6--0.9; it even inverts at small scale); state-space deviation ranks methods \emph{correctly}---there and at 7B on GSM8K/HumanEval (Appendix~\ref{app:merge7b}). In deep degradation, evaluate merges in state space.
\item \textbf{Transport dominates, and the cross-term is regenerated, not stored (\S\ref{sec:carried}, \S\ref{sec:erasure}).} An exact per-layer decomposition $\I_{\ell+1} = \G_\ell + T_\ell + M_\ell$ attributes $\sim$70\% of the flux to transport with late-network gain 1.08/block; erasure is then undone by propagation ($\rho = 0.99$, cosine 0.99) except near the output, where the durable causal effect concentrates.
\item \textbf{Cross-term erasure, and a model of it (\S\ref{sec:erasure}).} The first activation-level causal intervention on task-vector cross-terms: against five norm-matched controls (including steepest descent) and a dose ladder, erasure removes $\sim$19\% of expressed interference without damaging behavior, every other direction backfires or does nothing, and removable interference differs 14$\times$--337$\times$ between pairs whose magnitudes agree to within 15\%. A two-assumption model derives the measured angle law, retro-dicts the dose curve and depth profile, and passes a registered test of its own scaling prediction.
\item \textbf{Expression gating, and the audit of what the measure tracks (\S\ref{sec:gating}--\ref{sec:metrics}).} Instruction wrappers \emph{amplify} internal cross-term generation while collapsing its expression 20$\times$; erasure confirms the gate causally (13$\times$). The measure is decoupled from behaviour across four instruments, its cross-condition value is denominator-dominated, and naive bf16 generation is 75--90\% quantization roughness (\S\ref{sec:bf16}). Format-wrapped evaluations read the gate, not the interference.
\end{itemize}
Collectively, these results give the measure a causal anatomy and then withdraw the interpretation usually placed on it. All hypotheses, thresholds, and controls were preregistered in a versioned freeze log; falsified predictions, including two of three headline expectations, are reported as such---the mechanism above is what survived. Reduced replications at a second family (Llama-3.2-1B) and a larger scale (Qwen2.5-7B) reproduce all three pillars (Appendices~\ref{app:replication}, \ref{app:scale}).

\section{Related work}
\label{sec:related}

\begin{table}[t]
\centering\small
\caption{\textbf{No prior quantity is causal, directional, and format-aware; none explains the 2$\times$/20$\times$ structure} (empirical evaluation in \S\ref{sec:metrics}). Rows: CTL \citep{zhou2024ctl}, SurgeryV2 \citep{yang2024surgeryv2}, TSV \citep{gargiulo2025tsv}, NeuroMerging \citep{fang2025neuromerging}, APL \citep{kong2024apl}, RAIN-Merging \citep{rain2026merging}. Mag./Dir.: magnitude vs.\ direction quantity; Interv.: causal intervention on the quantity itself. CTL, parameter cosine, and SurgeryV2's representation bias are \emph{measured} in our conditions (\S\ref{sec:metrics}, Appendix~\ref{app:surgerybias}), not only cited.}
\label{tab:compare}
\setlength{\tabcolsep}{4.5pt}
\footnotesize
\begin{tabular}{lcccccc}
\toprule
 & Mag./Dir. & Interv. & Format & 2$\times$ pairs & 20$\times$ formats & Held-out \\
\midrule
CTL & mag. & -- & -- & no & no & -- \\
SurgeryV2 & mag. & -- & -- & no & no & -- \\
TSV & dir.\ (param.) & -- & -- & weak ($\rho{=}.25$) & no & weak \\
NeuroMerging & mag. & -- & -- & no & no & -- \\
APL & mag. & param. & -- & no & no & -- \\
RAIN-Merging & (behav.) & repair & \checkmark & -- & behav. & -- \\
\midrule
\textbf{Ours} & \textbf{dir.\ (activ.)} & \textbf{\checkmark\,(4 ctrl.)} & \textbf{\checkmark} & \textbf{14--337$\times$} & \textbf{13$\times$ gate} & $\rho{=}.65$ \\
\bottomrule
\end{tabular}
\end{table}

\paragraph{Task arithmetic and model merging.} Task vectors compose skills by weight addition \citep{ilharco2023task,ortiz2023tangent}; a large literature improves merging by resolving parameter-space conflicts---sign agreement \citep{yadav2023ties}, layer selection \citep{chen2025lata}, singular-vector overlap \citep{gargiulo2025tsv}, neuron-level decomposition \citep{fang2025neuromerging}---or by post-hoc repair of representation bias \citep{yang2024surgeryv2}; recent variants disentangle models \citep{ri2026resolving} or minimize input-subspace interference \citep{cheng2025wudi}, closed-form feature drift \citep{sun2025lot}, and singular-space inconsistency \citep{zhang2026dcmerge}. Task-level collapse is predicted correlationally by representational incompatibility, not parameter conflict \citep{cao2026collapse}; task information provably survives in merged weights \citep{wang2024tall}; mergeability predictors vary by method \citep{zhou2026mergeability}, degrade with overtraining \citep{horoi2025memorization}, and admit certified abstention \citep{gong2026twistedmerge}; and classification-time evaluation conflates representations with classifier mismatch \citep{kong2024rethink}. These methods and diagnoses \emph{presuppose} an account of where interference lives; our results test and revise it. ``Direction'' here is three claims---\emph{parameter-space} (TSV, DC-Merge), \emph{activation-space} (the orientation of $\I_\ell$; measured), \emph{causal} (what intervening does; tested)---and the third is load-bearing. APL \citep{kong2024apl} intervenes on parameters; ours on the cross-term itself.

\paragraph{Linearity of features under weight interpolation.} Cross-Task Linearity \citep[CTL;][]{zhou2024ctl} is the closest prior work: it observes that layerwise features of merged models are approximately linear combinations of the endpoint models' features. Our $\I_\ell$ \emph{is} the residual of that approximation, so our measurements refine theirs (in float32 even better than reported: cosines 0.93--0.98, \S\ref{sec:metrics}). The difference is what each work concludes from the smallness. CTL establishes that the layerwise cross-term is small; we establish that this small residual---specifically its orientation, not its magnitude---is where the functional interference causally lives, which magnitude-level analysis cannot see (Table~\ref{tab:compare}, \S\ref{sec:erasure}--\ref{sec:metrics}). CTL contains no intervention, no format factor, and no propagation analysis.

\paragraph{Format sensitivity of merged models.} RAIN-Merging \citep{rain2026merging} documents, at the behavioral level, that merging disrupts instruction-format handling and repairs it in the null space. Our expression-gating result is the mechanism-level counterpart---the wrapper closes an output gate on internally amplified interference---and neither result subsumes the other: they repair behavior; we locate what the behavior hides (\S\ref{sec:gating}).

\paragraph{Validity of activation patching.} Our interventions extend the locate--steer--improve programme of actionable mechanistic interpretability \citep{zhang2026locate} to merged models. Activation-level counterfactuals carry two known caveats: patched states may interact with unpatched components \citep{vaidyanathan2026mediators}, and interventions can push states off the natural distribution, recruiting pathways that never fire naturally \citep{grant2026divergent}. We claim no natural mediation: every effect is the response to a specified, norm-matched intervention, each control's directional overlap disclosed (\S\ref{sec:interventions}). At $\lambda{=}1$ the patched state is exactly $\bar h_\ell$, the additive reconstruction the estimand is defined against, and the displacement is 1.5--14\% of the state norm---at most $0.6$ of the model's own prompt-to-prompt spread; doses $\lambda > 1$ are flagged off-manifold. Appendix~\ref{app:validity} reports this divergence and the context-sensitivity checks \citet{grant2026divergent} ask for.

\section{The factorial ledger and cross-term erasure}
\label{sec:method}

The intuition first. Merging two task vectors and running a prompt gives four natural models---base, each single task, and the merge---and the \emph{interference} at any point in the computation is exactly what the merge has that the two single tasks do not add up to. Our method is nothing more than taking this four-way difference seriously at every layer, and then removing it surgically to see what changes.

\subsection{Setup}
\label{sec:setup}
We follow the training and measurement protocol of the companion measurement study \citep{paper1} exactly. Base model Qwen2.5-1.5B \citep{qwen25} (weights pinned to a single commit), six tasks (math, code, instruction, safety, summarization, translation \citep{cobbe2021gsm8k,taori2023alpaca,ji2024pku,narayan2018xsum,tiedemann2012opus}) finetuned as rank-16 LoRA \citep{hu2022lora} task vectors with response-only loss, three training seeds each. Task vectors are norm-matched to the global core-median norm before composition, and merged models $\theta_0 + \alpha\Delta_A + \beta\Delta_B$ are evaluated on a $6\times 6$ interior grid of $(\alpha,\beta)\in(0,1.2]^2$. Each prompt stratum contains 60 fixed prompts; all hidden-state quantities are read at the final prompt token. Our running example throughout is the \csafety{} merge evaluated on code prompts. Every confirmatory run executes on a single A100 with deterministic attention kernels and TF32 disabled; \S\ref{sec:validity} explains why this is not pedantry.

\subsection{Layerwise ledger: carried vs.\ generated}
\label{sec:ledger}
For a prompt $x$, let $h^{00}_\ell, h^{A0}_\ell, h^{0B}_\ell, h^{AB}_\ell$ be the residual state entering block $\ell$ under the four settings of Figure~\ref{fig:setup}. The \textbf{cumulative cross-term}
\begin{equation}
\I_\ell \;=\; h^{AB}_\ell - h^{A0}_\ell - h^{0B}_\ell + h^{00}_\ell
\end{equation}
is everything non-additive accumulated before block $\ell$. To separate what block $\ell$ \emph{adds} from what it \emph{transports}, we evaluate all four block functions at one \textbf{common reference state} $\bar h_\ell = h^{A0}_\ell + h^{0B}_\ell - h^{00}_\ell$ (the additive reconstruction) and take the same four-corner difference:
\begin{equation}
\G_\ell \;=\; F^{AB}_\ell(\bar h_\ell) - F^{A0}_\ell(\bar h_\ell) - F^{0B}_\ell(\bar h_\ell) + F^{00}_\ell(\bar h_\ell).
\end{equation}
Intuitively, $\G_\ell$ is the cross-term block $\ell$ would create from scratch on interference-free input; the exact ledger $\I_{\ell+1} = \G_\ell + P_\ell$ then attributes the remainder $P_\ell$ to transport of what was already carried (we do not interpret $P_\ell$ beyond this accounting identity). Two structural identities hold exactly and serve as built-in sanity checks: $\I_1 = 0$ (LoRA leaves the embedding untouched) and $\I_2 = \G_1$; both are verified to machine zero in every run.

\subsection{Output-side estimands}
\label{sec:estimands}
Following \citet{paper1}, expressed interference on a prompt batch is the interaction ratio $\Rint$: the mean Jensen--Shannon divergence between the merged model's next-token distribution and its no-interaction additive prediction (logit-space $\ell^{A0}+\ell^{0B}-\ell^{00}$), normalized by the mean main-effect JSD---a ratio of means, never a mean of ratios. The causal effect of an intervention is
\begin{equation}
\Dint \;=\; \Rint_{\text{original}} - \Rint_{\text{patched}},
\end{equation}
with the \emph{unpatched} main-effect denominator frozen in both terms, so $\Dint$ moves for exactly one reason. Preregistered contrasts are paired differences of these summaries across conditions sharing prompts, evaluated with prompt-paired bootstrap CIs and 3/3 seed-direction agreement, seeds being the outer replication unit.

\subsection{Cross-term erasure and its controls}
\label{sec:interventions}
The intervention replaces the merged model's residual state at a fixed boundary with $h^{AB}_\ell - \lambda \I_\ell$: at $\lambda=1$, the carried cross-term is exactly erased and the state is restored to the additive reconstruction; the ladder $\lambda\in\{0, 0.5, 1, 1.5, 2\}$ probes dose dependence. The claim ``the \emph{direction} of $\I_\ell$ is what matters'' is only as strong as its controls, so every control preserves $\lVert \I_\ell\rVert$ per prompt while destroying exactly one structural property, with each control's overlap $\cos(\delta, -\I_\ell)$ reported so direction-mediation is checked, not assumed:
\begin{itemize}\itemsep1pt
\item \textbf{wrong-pair}: the additive state borrowing the second axis from the \emph{other} task pair (kills: ``any task-structured direction works'');
\item \textbf{coefficient-mismatch}: the $(+1,-1,+1)$ combination in the same affine span (kills: ``any span-consistent displacement works'');
\item \textbf{cross-prompt}: the additive state of a \emph{donor} prompt (fixed length-binned derangement), applied as a full dose ladder (kills: ``prompt identity is irrelevant'');
\item \textbf{distance-matched random} (kills: ``perturbation per se helps'').
\end{itemize}
Interventions run in float32; damage is audited both at the first token (top-1 agreement, entropy shift, JSD from unpatched) and over prefill-patched greedy generation on both endpoint strata.

\subsection{Numerical validity, preregistration, and why both were load-bearing}
\label{sec:validity}
The four-corner difference is a catastrophic cancellation---the four paths differ at $O(\alpha)$ while $\I_\ell$ is $O(\alpha\beta)$---so every estimator in this paper carries its own noise floor and gates: (i) a \emph{determinism null} (four forwards at identical coefficients must agree bitwise; enforced, not assumed); (ii) a \emph{propagated-ULP floor} (half-ULP representation jitter on the four stored states, propagated in float64 through the cancellation; any layer below its floor is marked uninterpretable); (iii) finite-difference probes carry step-halving convergence checks; (iv) the exact identities of \S\ref{sec:ledger}. These gates are not decoration: before any conclusion depended on them they caught an all-bf16 mixed-derivative estimate that was pure noise, a floor computation silently zeroed by float32 rounding, and ultimately the \S\ref{sec:bf16} finding itself.

All hypotheses, criteria, layer positions, dose ladders and control constructions were frozen---in a versioned freeze log with a full change record---before the corresponding data existed; falsified predictions, including two of three headline expectations, are reported as falsified with their prior confidences. Verdicts, immutable per-cell results and the full freeze chain are released.

\section{Transport dominates generation---and magnitude is not the signal}
\label{sec:carried}

We begin with the question the field's metrics implicitly answer: \emph{how much} cross-term does each layer produce, and does that quantity carry the functional differences?

Measured naively at training precision, local generation looks like a universal constant---a result that dissolves under audit, and instructively so. Across all twelve (pair $\times$ stratum) conditions of our main sweep---1{,}296 cells spanning two task pairs, six prompt strata, three seeds, and the full composition grid---the bf16 full-depth generation summary AUC($\G$) occupies a band of 0.0101--0.0116 (per-pair ratios 1.02--1.15$\times$), including the geometrically most conflicting and most orthogonal pairs. The frozen precision audit (\S\ref{sec:bf16}) attributes this uniformity to a task-independent quantization floor. Recomputed with float32 blocks over the full grid (36 cells), the real picture is different: the fp32 band spans 3.4$\times$ (0.0012--0.0040, falsifying our own frozen $<$2$\times$ expectation), the pair ratio is 1.81--1.90$\times$---\emph{coarsely tracking} the 2$\times$ output difference, as every magnitude quantity in this paper turns out to (\S\ref{sec:metrics})---and generation under instruction wrappers is 1.7--1.8$\times$ \emph{larger} than raw, 3/3 seeds.

The output side sets the standard these numbers must meet. The two pairs differ by roughly 2$\times$ in expressed interference (family scores 10.33\% vs.\ 5.31\%) and---causally, by erasure---14$\times$--337$\times$ in removable interference (\S\ref{sec:erasure}); the same prompts differ by roughly 20$\times$ across formats ($+6.91$pp raw vs.\ $+0.34$pp wrapped) \citep{paper1}, in the \emph{opposite} direction from generation. Generation magnitude is at best a coarse correlate of pair structure and is inverted on the format axis; we call the assumption that it carries the functional differences the \textbf{magnitude fallacy} (Figure~\ref{fig:one}a).

\subsection{A cautionary result: naive bf16 estimates of $G$ are mostly round-off}
\label{sec:bf16}
Before interpreting the uniform band, we report what it is made of. Recomputing $\G_\ell$ on the \emph{same} reference state with the block cast to float32 shrinks it to 14--23\% of its bfloat16 value, with directions nearly uncorrelated across precisions (median cosine 0.15--0.24): 75--90\% of the naively measured ``local generation'' is bfloat16 weight-rounding roughness---and rounding roughness is task-independent, which is precisely \emph{why} the band is universal. We flag this as a standalone caution: activation-difference analyses run at training precision can be dominated by quantization roughness while passing casual sanity checks (cf.\ the BF16 training--inference mismatch in RL fine-tuning, cured by FP16; \citealp{qi2025fp16}). Paired \emph{contrasts} survive---the rounding floor is condition-independent and cancels in differences---and all causal results are computed in float32.

\subsection{Transport dominates: an exact decomposition, plus the injection bounds}
\label{sec:amplify}
\begin{figure}[t]
\centering
\includegraphics[width=0.78\textwidth]{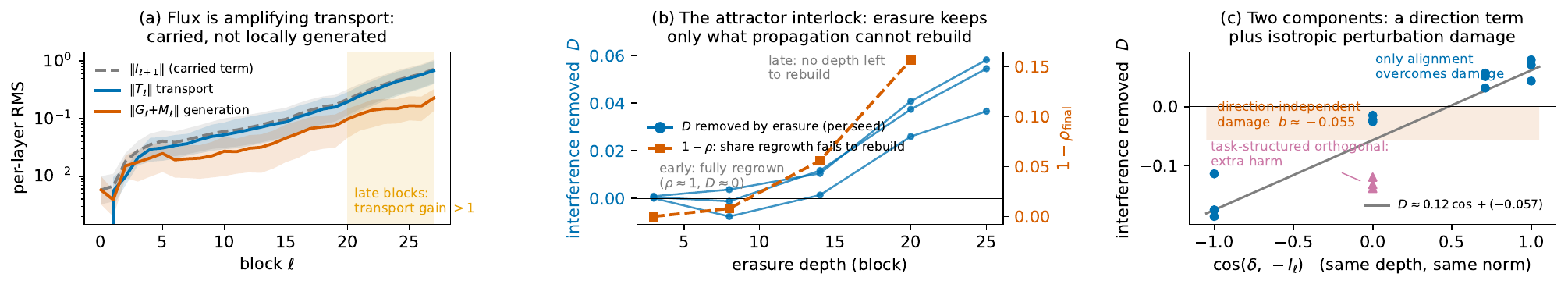}
\vspace{-5mm}
\caption{\textbf{The mechanism in three steps: the cross-term is transported and amplified (a), regrown when erased early (b), and removable only along its own direction (c).} (a) Exact flux decomposition (2 pairs $\times$ 3 seeds, float32; bands span cells): transport $T_\ell$ tracks the carried term and dominates generation; shaded late blocks: gain $>1$. (b) What regrowth fails to rebuild ($1-\rho$, right axis) and what erasure durably removes ($\Dint$, per seed) rise together with erasure depth. (c) At fixed depth and norm, $\Dint$ tracks $\cos(\delta, -\I_\ell)$ (points: seeds; line: mean fit): a direction term over a damage intercept; orthogonal arms \emph{worsen} interference (Appendix~\ref{app:orthogonal}).}
\label{fig:mech}
\vspace{-2mm}
\end{figure}

The accounting identity $\I_{\ell+1} = \G_\ell + P_\ell$ leaves $P_\ell$ uninterpreted, and rightly so: it mixes transport of the existing term with state--parameter coupling. We therefore decompose it \emph{exactly}. Evaluating all four blocks (float32) both at the common reference $\bar h_\ell$ and at the four actual path states regroups the actual-state four-corner term, bitwise, as
\begin{equation}
\I_{\ell+1} \;=\; \G_\ell \;+\; \underbrace{F^{AB}_\ell(h^{AB}_\ell) - F^{AB}_\ell(\bar h_\ell)}_{T_\ell\ \text{(transport of the existing }\I_\ell\text{)}} \;+\; M_\ell,
\end{equation}
where $M_\ell$ bundles the marginal-path state mismatches (the state--parameter coupling channel); $\G_\ell + M_\ell$ is precisely the one-step regeneration a block produces when $\I_\ell$ has just been erased. Over the 36 fp32 cells: transport carries $\sim$69\% of the layerwise flux (regeneration share $0.31$, 3/3 seeds), late-network transport exceeds regeneration by 2.3$\times$, and \emph{transport alone amplifies}---$\lVert T_\ell\rVert / \lVert \I_\ell\rVert = 1.08$ per late block, 3/3 seeds (Appendix~\ref{app:decomposition}; Figure~\ref{fig:mech}a). A replication on Llama-3.2-1B reproduces both: transport share $65\%$, late gain $1.14$, 3/3 seeds---transport dominance is a family invariant. The coarser arithmetic exclusion agrees: if layers contributed independent increments of the measured $\lVert\G_\ell\rVert$, cumulative growth is bounded by $0.088$ (incoherent) and $0.369$ (perfectly aligned), against an observed $\lVert \I_{28}\rVert = 0.974$---2.6$\times$ even the coherent bound. Consistently, smooth local generation is small and flat---mixed-derivative stencils at five centers of the composition rectangle are level to within 1.4$\times$---in agreement with CTL's approximate-linearity observation \citep{zhou2024ctl}, while relocating everything that observation leaves unexplained into propagation.

\emph{Refuted conjecture:} we predicted (55\% prior) that mixed-derivative mass concentrates off-diagonal; measured edge-to-diagonal ratios of 1.005--1.010 (max 1.37, zero blocks above the preregistered 2$\times$ threshold) falsify it.

\section{Orientation is load-bearing: cross-term erasure}
\label{sec:erasure}

Section~\ref{sec:carried} shows magnitude cannot carry the differences. This section shows what does---causally. We intervene at five fixed depths with \erasure{} and the four norm-matched structural controls of \S\ref{sec:interventions}, with all three predictions registered before implementation.

\subsection{Erasure works; every norm-matched control fails or backfires}
At $\lambda=1$ on code prompts (\csafety{}, mean over five depths and three grid points), erasure removes roughly 19\% of expressed interference---a depth-average the regrowth analysis below decomposes---and the specificity contrasts are positive against \emph{every} control, 3/3 seeds (Table~\ref{tab:kt2}). The wrong-pair and coefficient-mismatch patches in fact have \emph{negative} $\Dint$: an equally large, equally structured displacement in the wrong direction makes interference worse. Conditioning on each control's overlap $\cos(\delta, -\I_\ell)$, the effect is proportional to the displacement's projection on the erasure direction (e.g., at depth 14: correct $\cos{=}1.00, \Dint{=}+0.028$; wrong-pair $\cos{=}0.25, \Dint{=}+0.007$), so we claim \emph{cross-term-direction specificity}---the direction is the mediator-facing variable---rather than task-identity specificity beyond it.

\subsection{Dose--response dissociates direction from displacement}
\looseness=-1 Along the correct direction, $\Dint(\lambda)$ rises and saturates near $\lambda\approx1$; along the norm-matched cross-prompt direction it \emph{worsens monotonically} (Figure~\ref{fig:one}b). No account in which perturbation per se helps or hurts survives both curves: sign is decided by direction alone, and the correct direction completes near $\lambda \approx 1$, exactly where the cross-term is fully erased.

\subsection{Matched magnitudes, $14\times$--$337\times$ apart in removable interference}
\looseness=-1 On the very code prompts where the two merges' local generation differs by at most 1.9$\times$ in float32 (\S\ref{sec:carried}), erasure removes substantial interference from \csafety{} and almost nothing from \cmath{} (Table~\ref{tab:kt2}, last row), giving pair ratios of $13.9/337/29.5$ (the middle value rides a near-zero denominator; in \emph{unnormalized} JSD the reductions differ by $18$--$52\times$). The pairs are matched on local generation but \emph{not} on the carried term---late-layer $\lVert \I \rVert$ differs 2.4$\times$ (0.65 vs.\ 0.27), disclosed because \S\ref{sec:theory} says that is what matters---yet scale cannot produce the gap: at depth 25 erasure removes 56\% of expressed interference from \csafety{} and $-8\%$ from \cmath{}, a change of \emph{sign}. The paper's central quantitative contrast: the pair difference lives in the orientation of the carried cross-term, far beyond what its size can explain.

Removal is not destruction. Erasure at $\lambda=1$ leaves first-token behavior almost untouched: top-1 agreement 0.94--0.99, patched-versus-unpatched JSD (0.003) two orders below the main effect, and 92--99\% of continuation tokens preserved.

\subsection{Erased early, re-created: the cross-term is not stored}
Propagation undoes erasure except where depth runs out (cf.\ emergent self-repair, \citealp{mcgrath2023hydra,rushing2024selfrepair}; distinguished in Appendix~\ref{app:regrowth}). If interference were \emph{stored} in the cross-term, erasing it should remove it for good. A follow-up measurement (window and priors fixed before data; Appendix~\ref{app:regrowth}) tracked the cross-term downstream of a $\lambda{=}1$ erasure, expecting partial regrowth ($\rho = \lVert \I^{\mathrm{post}} \rVert / \lVert \I^{\mathrm{orig}} \rVert \in (0.3, 0.95)$ at the output). The window was \emph{falsified from above}: erase at depth 3 or 8 and the network rebuilds the cross-term essentially exactly---final-boundary $\rho = 1.00/0.99$, direction cosine to the original cross-term $0.99$ (the frozen direction-memory prediction, $\cos > 0.5$, held with room to spare). Regrowth only falls off when depth runs out: $\rho = 0.94$ erasing at 14, $0.84$ at 20. And this profile interlocks with the causal effect: decomposing $\Dint$ by patch depth, early erasures achieve nothing ($\Dint \approx 0$ at depths 3 and 8, 3/3 seeds) while late erasures carry the entire effect ($+0.026$ to $+0.058$ at depths 20--25)---removable interference is exactly what the remaining depth cannot rebuild (Figure~\ref{fig:mech}b). The reading we draw (formed after these results): the carried cross-term is \emph{continuously re-established from the untouched marginal paths} rather than stored. What returns is the cross-term coordinate, not the state---a matched-displacement control finds no restoring force specific to this direction, and the perturbed trajectory never rejoins the unpatched one (Appendix~\ref{app:recovery}). This sharpens \S\ref{sec:amplify}'s exclusion into an interventional statement---propagation does not merely amplify the carried term, it actively reconstructs it in norm and direction---and yields a design corollary: erasure-style mitigation must act late, where reconstruction cannot catch up.

\subsection{Orthogonal controls: a direction term over a damage intercept}
Target and estimand share the additive counterfactual, so the reduction might reflect \emph{any} displacement toward the additive neighbourhood---a generic contraction---not direction. We test this by holding depth and norm fixed and varying \emph{only} the angle to $-\I_\ell$ (Appendix~\ref{app:orthogonal}; Figure~\ref{fig:mech}c): displacements orthogonal to $\I_\ell$---random, and a task-structured one---do \emph{not} reduce $\Rint$ but increase it ($\Dint = -0.015$ to $-0.139$), while $\Dint$ varies monotonically with $\cos(\delta, -\I_\ell)$ across five angles (3/3 seeds). A generic-contraction account predicts the opposite for the orthogonal arms, and is excluded---as it is in state space, where random orthogonal displacements are diluted twice as strongly as structured ones (Appendix~\ref{app:recovery}). A norm-matched steepest-descent arm \emph{backfires} ($\Dint$ $-0.17$ to $-0.65$, 9/9) while $-\I_\ell$---nearly orthogonal to it ($\cos \le 0.12$)---helps: not loss-alignment but the low-gain direction the quadratic model favours (Appendix~\ref{app:orthogonal}). The sweep also decomposes the effect: $\Dint \approx a\cos(\delta,-\I_\ell) + b$ with slope $a = 0.09$--$0.14$ and \emph{negative} intercept $b \approx -0.055$ ($R^2$ 0.64--0.90)---a direction-independent damage term that only alignment overcomes, derived below. The depth profile agrees: a construction-driven effect would appear at every depth, yet $\Dint \approx 0$ at depths 3--8, in proportion to $1-\rho$.

The coupling has a falsifiable consequence, and it fails. If moving the state onto $\bar h_\ell$ guaranteed moving the logits onto the additive prediction, erasure would help every pair at every depth. It does not, because $F(\bar h_\ell) \ne a$: the difference is the regeneration $g$, so the experiment measures $\lVert g\rVert^2/\lVert u_0\rVert^2$---a dynamical quantity, not a definitional one. For \cmath{} at depth 25 exact erasure \emph{increases} expressed non-additivity in two of three seeds ($+0.0020/-0.0051/-0.0051$). A construction artifact cannot change sign. (Post hoc on frozen data.)

\looseness=-1\emph{Statistical basis, disclosed:} the frozen verdict rests on 3/3 seed-direction consistency; a per-prompt re-run---reproducing every immutable scalar bitwise---supplied the bootstrap, and 27 of 28 intervals exclude zero, the exception being the frozen PK2-2 expectation itself (Appendix~\ref{app:prereg}). A two-level seed$\times$prompt bootstrap agrees: $\Dint_{\mathrm{correct}} = +0.018$ $[+0.010, +0.031]$, contrast vs.\ wrong-pair $+0.033$ $[+0.021, +0.049]$.

\begin{table}[t]
\centering\footnotesize
\setlength{\tabcolsep}{5pt}
\caption{\textbf{Cross-term erasure is direction-specific: the correct direction removes interference, every norm-matched control fails or backfires} ($\lambda{=}1$, code prompts, \csafety{}, mean over 5 fixed depths $\times$ 3 grid points; per-seed values). Brackets: 95\% prompt-bootstrap CIs (5{,}000 paired draws/seed) from the bitwise-reproducing per-prompt re-run; all 15 exclude zero.}
\label{tab:kt2}\vspace{-1mm}
\begin{tabular}{lrrr}
\toprule
 & seed 42 & seed 123 & seed 456 \\
\midrule
$\Dint$(correct) & $+0.0301$ {\tiny$[.021,.042]$} & $+0.0289$ {\tiny$[.023,.037]$} & $+0.0160$ {\tiny$[.013,.020]$} \\
$\Dint$(correct)$-\Dint$(wrong-pair) & $+0.0556$ {\tiny$[.042,.074]$} & $+0.0587$ {\tiny$[.044,.077]$} & $+0.0487$ {\tiny$[.040,.059]$} \\
$\Dint$(correct)$-\Dint$(coeff-mismatch) & $+0.0699$ {\tiny$[.053,.091]$} & $+0.0719$ {\tiny$[.056,.091]$} & $+0.0583$ {\tiny$[.048,.069]$} \\
$\Dint$(correct)$-\Dint$(cross-prompt) & $+0.0460$ {\tiny$[.038,.055]$} & $+0.0391$ {\tiny$[.032,.047]$} & $+0.0257$ {\tiny$[.021,.031]$} \\
$\Dint$(correct)$-\Dint$(random) & $+0.0375$ {\tiny$[.027,.050]$} & $+0.0355$ {\tiny$[.029,.044]$} & $+0.0206$ {\tiny$[.017,.025]$} \\
\midrule
$\Dint$(correct), \cmath{} & $+0.0022$ & $+0.0001$ & $+0.0005$ \\
\bottomrule
\end{tabular}
\end{table}

\subsection{A minimal model that predicts the law}
\label{sec:theory}
\looseness=-1 Two assumptions give it. (A1) The map $F$ from the patched depth to the logits is differentiable at $\bar h_\ell$ with Jacobian $J$, and the displacement---measured at 1.5--14\% of the state norm---is small enough for a first-order expansion. (A2) The estimand is the quadratic form $R = \lVert u\rVert^2_M$, the second-order expansion of the JSD. With $g = F(\bar h_\ell) - a$ the interaction the remaining depth \emph{regenerates}, $u(\delta) \approx g + J(\I_\ell + \delta)$, and with $u_0 = g + J\I_\ell$ the effect is exactly $\Dint(\delta) = -2\langle u_0, J\delta\rangle_M - \lVert J\delta\rVert^2_M$. The first term pays only for alignment with $-\I_\ell$; the second is negative for \emph{every} displacement, so that intercept is structural. The model retro-dicts three frozen measurements (labelled as such). It forces the dose curve to be a two-parameter parabola through the origin---the measured ladder fits at $R^2 = 0.99$--$1.00$, peaking above $\lambda{=}1$ exactly when regeneration aligns with transport---the independently measured cosine-$0.99$ direction memory; the dose and angle fits estimate the same $\langle u_0, J\I_\ell\rangle$ from data frozen months apart and agree to 7--13\% with no shared parameter; and since the effective quantity is $J\I_\ell$, not $\lVert \I_\ell\rVert$, the effect must vanish wherever propagation rebuilds the cross-term. Its one untested prediction---$a \propto s$ but $\lvert b\rvert \propto s^2$---we froze and ran: $p_a = 1.03/1.01/1.05$ against a predicted 1, while $p_b = 1.75/1.63/1.71$ sits inside its frozen window but below the predicted 2, which JSD's bounded range would produce---flagged untested (Appendix~\ref{app:theory}).

\section{Expression gating: formats hide interference}
\label{sec:gating}

\looseness=-1 The magnitude fallacy has a second casualty: the assumption that what an evaluation \emph{sees} tracks what the model \emph{carries}. The format bridge tests it---the same 60 public code prompts, raw and wrapped in an Alpaca template---where the companion study established a 20$\times$ collapse of expressed interference ($+6.91$pp$\,\to\,+0.34$pp) \citep{paper1}.

\looseness=-1 Instruction wrappers amplify internal interference while collapsing its expression. We preregistered the natural expectation---internal generation should shrink with the output---and the data reversed it: wrapping \emph{increases} local generation: $C_{\text{format}} = -7.5\times10^{-4}$ (\csafety) and $-6.1\times10^{-4}$ (\cmath), CIs excluding zero, 3/3 seeds, companion agreeing in sign. The template is in-distribution for both adapters: both axes activate and their mixture grows, while the pinned output distribution never surfaces it. We call this \emph{expression gating}---a readout-level dissociation, not an internal switch.

\looseness=-1 Erasure confirms the gate causally: where nothing is expressed, little \emph{relative} interference is removable. If the wrapper merely \emph{masked} noise, erasure would work equally under both formats; instead the identical $\lambda{=}1$ erasure yields $\Dint_{\text{raw}} = +0.0286/+0.0370/+0.0174$ but $\Dint_{\text{wrapped}} = +0.0031/+0.0026/+0.0005$---a 13$\times$ mean ratio, 3/3 (Figure~\ref{fig:one}c). At 7B the gate is fully closed: $\Dint_{\text{wrapped}}$ indistinguishable from zero, all seeds (Appendix~\ref{app:scale}).

\looseness=-1 The gate drowns the interaction rather than shrinking it: the collapse is denominator-driven: the wrapper inflates the main-effect JSD $\sim$13$\times$ (0.025$\to$0.32) while the \emph{absolute} interaction JSD does not shrink (raw/wrapped $0.4$--$1.0\times$), and the absolute JSD removed by erasure is format-comparable. The gate is \emph{drowning}, not destruction: the template pins the output distribution, and a similar-sized interaction becomes negligible against it. The structure generalizes across two further instruction templates (denominator 11--35$\times$, $\Rint$ down 18/18) and is absent for a length-matched non-instruction prefix---and, out of sample, for an untrained ChatML serialization, which patterns with the control (Appendix~\ref{app:wrappers}).

\looseness=-1 The methods that work make the measure worse---and under a shared reference it says nothing. We audited four weight-space methods---task arithmetic, trimming, TIES, DARE---on shared task vectors (Appendix~\ref{app:merge}). No method reduces $\Rint$ (0/3 seeds, frozen expectation); TIES roughly \emph{doubles} it. Under a common counterfactual it carries no method information: four-method range $0.006$--$0.015$ against behavioural spreads of $0.6$--$0.9$. Per-method conventions even rank the best method last in the small-scale cell (TIES: most interference, best behaviour)---an inversion that does not reproduce at 7B (\S\ref{sec:metrics}). Either way, the output-side ratio cannot rank methods.

\looseness=-1 Where merging collapses, the cross-term is a bystander. A fourth behavioural attempt ran erasure in the collapse cell: erasing the entire cross-term leaves the collapsed code score \emph{exactly} unchanged (final boundary, 3/3), while at this scale the equal-norm $+\I$ arm destroys it ($0.40\to0.00$); no boundary repairs (0/3, preregistered). A small-dose ladder excludes cancellation (flat correct-arm dose curve; quadratic orthogonal degradation). A second collapse parameterization (7B, real benchmarks) reproduces the null repair (recovery $\le$4\%, 3/3) but not the destructive doubling---its displacements are half as large. The carrier has its own dose-response: $\alpha$ moves behaviour; erasure---even persistent, realizing the additive counterfactual online and finding it \emph{still collapsed}---does not (Appendix~\ref{app:repair}). The residue can damage; removing it cannot repair: collapse rides in the marginal displacements---the \emph{carrier--bystander split} (Appendix~\ref{app:repair}).

\looseness=-1 Implication for evaluation practice: A merged model can carry \emph{more} interference precisely when a format-wrapped evaluation reports it \emph{cleanest}. The gate is task-count dependent: six task vectors rather than two raise $\Rint$ under the same template from $0.003$ to $0.82$ (Appendix~\ref{app:degradation})---a property of two-vector merges, not a law. Within that scope, wrapped-only assessment reads the gate---the mechanism-level counterpart of RAIN-Merging's format fragility \citep{rain2026merging}. We make no claims about safety \emph{behaviors}.

\section{What existing metrics see---and don't}
\label{sec:metrics}

Can the metrics the field already uses \emph{predict} interference for unseen pairs, and \emph{explain} the causal structure of \S\ref{sec:erasure}--\ref{sec:gating}? Across 15 pairs of six tasks (6 calibration, 9 held-out) we compare CTL linearity and residual, parameter cosine, direction features, and cumulative magnitude against measured interference, with only feature \emph{signs} fitted in calibration.

\looseness=-1 Linearity per se is uniformly high and separates nothing. CTL's central observation replicates \emph{more strongly than reported} (cosines 0.93--0.98, every family), but precisely because it is uniform it cannot explain why one pair interferes at 2$\times$ another or a wrapper changes expression 20$\times$; parameter cosine fares little better ($\rho = +0.25$ held out).

\looseness=-1 Everything in the magnitude family sees coarse pair structure and nothing else. SurgeryV2's bias (36 cells, Appendix~\ref{app:surgerybias}) orders the pair contrast (2.9$\times$, 3/3) but on the format bridge moves between $-1\%$ and $+34\%$ while expression collapses $\sim$95\%. Cumulative $\lVert \I \rVert$ predicts held-out pair ordering at $\rho = +0.63$--$0.65$ (3/3 seeds, 8/8 preregistered comparisons; directional, not significant), which \S\ref{sec:carried} explains: propagation gain is pair-dependent. Both stay blind on formats, where every magnitude points the \emph{wrong way}---axis-specifically: \emph{absolute} interaction does track coefficient-driven degradation (11/15 sweeps), while on the method axis absolute and normalized are equally blind (Appendix~\ref{app:merge7b}); \emph{our own estimand belongs on this list}: it too fails to rank merging methods (\S\ref{sec:gating}).

\looseness=-1 The family splits in state space. The carrier--bystander split, in metric form: in the collapse cell, SurgeryV2-style bias and a CTL-style state residual---the \emph{same} relative non-additivity as $\Rint$, read before the readout---rank the methods \emph{correctly} ($-0.8$ to $-1.0$, 3/3 seeds) while $\Rint$ ranks them backwards: the inversion lives in the readout map, where the theory's $J$ sits (\S\ref{sec:theory}). A 7B gate (GSM8K-test, executed HumanEval) confirms the state-space ranking ($-1.0$, both endpoints) and flattens $\Rint$'s range to $0.06$ in the deepest cell (Appendix~\ref{app:merge7b})---the interventional counterpart of the correlational finding of \citet{cao2026collapse}.

\looseness=-1 Our own registered direction features failed: the best ($\rho = +0.62$) is magnitude in disguise ($\propto 1/\lVert \I \rVert$ via its logit-lens normalization); direction's causal role rests on intervention.

\subsection{What to use instead}
The audit yields a three-line recipe. (1) For coarse pair screening, any magnitude quantity---cumulative $\lVert \I \rVert$, SurgeryV2 bias, fp32 generation---is a cheap screen ($\rho \approx 0.65$ held out), and nothing finer: do not read differences below its $\sim$2$\times$ resolution. (2) For evaluating merges in the degraded regime, read the state, not the readout: SurgeryV2-style bias or the CTL-style state residual at a late boundary, computable from forward passes the evaluation already runs, rank methods correctly at both scales where $\Rint$ carries no signal. (3) Report interference conditional on serialization: raw prompts see what tuning-template harnesses hide, and no magnitude read under one format transfers to another (\S\ref{sec:gating}).

\section{Discussion, limitations, and conclusion}
\label{sec:conclusion}

Effects here are responses to norm-matched interventions---direction-mediation under intervention, not natural mediation; within a merge, direction governs the response, and across merges damage tracks displacement magnitude (\S\ref{sec:gating}). \looseness=-1 Two further model points reproduce the structure. Llama-3.2-1B and Qwen2.5-7B replications (Appendices~\ref{app:replication}, \ref{app:scale}) give direction-specific erasure (all bootstrap intervals exclude zero; pair separations 2.6--5.5$\times$, 6.9--22.9$\times$); one $\G$-band gate missed by 0.1\% (reported failed). Gating strength is a family trait ($\approx$2$\times$ Llama, total at 7B); the 13$\times$/20$\times$ figures are scoped to Qwen2.5-1.5B: mechanism invariant; every \emph{magnitude} not.

\looseness=-1\textbf{Limitations, each with its direction.} (i) \emph{Three model points, one adapter parameterization} (rank-16 LoRA; Appendix~\ref{app:merge}). (ii) \emph{CIs are within-seed}; seeds are the outer unit; one 3/3 gate has $p{=}1/8$---the ledger's weight is breadth. (iii) \emph{Controlled prompts are diagnostic instruments}; behaviour-level validation failed three times and a fourth, collapse-regime attempt found erasure inert: it controls output non-additivity, not behaviour (Appendices~\ref{app:behavior}, \ref{app:repair}). A multi-position check excludes readout position (Appendix~\ref{app:behavior}). (iv) \emph{The 7B collapse criterion fires in 1/3 seeds} (others 23--25\%).

\looseness=-1\textbf{Conclusion.} What survived is not what we set out to confirm: the cross-term has a causal anatomy---transported, amplified, regrown, direction-selective, a bystander in collapse---none of which licenses the measure; state-space deviation does. Before a quantity can diagnose a merge, someone has to ask what it sees.

\subsubsection*{Reproducibility statement}
All hypotheses, thresholds, layer positions, dose ladders, and control constructions were frozen before the corresponding data existed, in a versioned preregistration (six freeze versions, each superseding change logged); all fourteen predictions and their outcomes, three verdict documents, 1{,}296+90+54 immutable per-cell results plus the 54 per-prompt intervention records and their bootstrap-CI report, the second-family replication preregistration, verdict, and its 54+54 cells, the frozen regrowth follow-up and its 18 cells, the frozen SurgeryV2-bias comparison and its 36 cells, the frozen scale point and its 54+54 cells, the frozen transport-decomposition/fp32-grid experiment and its 36 cells, the intervention-divergence measurement and its 6 cells, the frozen merging-method audit with its metric-family, common-counterfactual, and 7B real-benchmark replications (36+36+12+51 cells and four verdicts), the frozen collapse-regime behavioural loop with its small-dose ladder and 7B second-parameterization replication (78+54+63 cells, three verdicts), the multi-position readout check (3$\times$60 prompts), the frozen steepest-descent control (27 cells) and three-layer reanalysis with their verdicts, the numerical-validity audits (determinism nulls, ULP floors, bitwise equivalence audits), the exact software stack, and the pinned base-model commit are released with the code. Every figure is generated by a script whose numbers come from the immutable cell records or the frozen verdict tables, never retyped.

\subsubsection*{Statement on AI usage}
Large language models were used, under continuous author direction and review, to assist with experiment-code implementation, data-analysis scripting, and manuscript drafting and editing. All hypotheses, preregistered predictions, pass/fail criteria, and verdicts were specified or approved by the authors; every reported number derives from immutable experiment records released with the code, and all figures are regenerated from those records by released scripts. No LLM was used to generate, impute, or select experimental data.

\bibliography{references}
\bibliographystyle{plainnat}

\appendix

\section{Full estimator and aggregation definitions}
\label{app:estimators}
All definitions below are transcribed from the frozen design document (\S\ref{app:prereg}); none was altered after data existed.

\paragraph{Four weight-state paths.}
For task vectors $\Delta_A, \Delta_B$ and coefficients $\alpha,\beta$, we instantiate four weight states $M_{00}$, $M_{\alpha 0}$, $M_{0\beta}$, and $M_{\alpha\beta}$ and, for a prompt $x$, record the residual state at the input of block $\ell$ under each: $h^{00}_\ell$, $h^{\alpha 0}_\ell$, $h^{0\beta}_\ell$, and $h^{\alpha\beta}_\ell$. All hidden analyses use the \emph{last prompt token} only, aligning with the companion study's first-generated-token logit proxy and avoiding position-averaging bias between raw and wrapped sequences of different lengths.

\paragraph{Cumulative cross-term (descriptive only).}
$\I_\ell = h^{\alpha\beta}_\ell - h^{\alpha 0}_\ell - h^{0\beta}_\ell + h^{00}_\ell$. Because $\I_\ell$ contains every nonlinearity produced and propagated by blocks $1..\ell{-}1$, depth curves of $\I_\ell$ are frozen as descriptive: they cannot by themselves locate where cross-terms are \emph{generated}.

\paragraph{Common-state local generation (KT1 primary object).}
With $F^{ab}_\ell$ the full block-$\ell$ map (residual update included) of weight state $M_{ab}$, and the additive reference $\bar h_\ell = h^{\alpha 0}_\ell + h^{0\beta}_\ell - h^{00}_\ell$,
\[
\G_\ell(\bar h_\ell) = F^{\alpha\beta}_\ell(\bar h_\ell) - F^{\alpha 0}_\ell(\bar h_\ell) - F^{0\beta}_\ell(\bar h_\ell) + F^{00}_\ell(\bar h_\ell).
\]
Because all four block paths receive the identical input, $\G_\ell$ isolates the four-corner term contributed by the combined weights at the current reference state. The exact ledger $\I_{\ell+1} = \G_\ell + P_\ell$ defines the \emph{state-conditioned remainder} $P_\ell$, which is frozen as \emph{not} interpretable as pure propagation (it mixes propagation of existing interaction, reference-state mismatch, state--parameter coupling, and higher-order block nonlinearity).

\paragraph{Profiles, AUCs, and companions.}
The primary layer profile is $g^{\mathrm{RMS}}_\ell = \mathrm{RMS}(\G_\ell)$ with full-depth AUC as the headline summary (early-depth AUC over normalized depth $[0,0.5]$ is a frozen secondary). The functionally weighted companion applies $\phi(z) = W_U\,\mathrm{FinalNorm}(z)$ to each corner before differencing; untuned logit-lens is confirmatory only at normalized depth $d \ge 0.5$, descriptive earlier. The two frozen contrasts are $C_{\text{pair}} = \mathrm{AUC}(\G;\text{\csafety}) - \mathrm{AUC}(\G;\text{\cmath})$ and $C_{\text{format}} = \mathrm{AUC}(\G;\text{raw}) - \mathrm{AUC}(\G;\text{wrapped})$.

\paragraph{Output estimand and intervention effect.}
$\Rint$ is the companion study's ratio-of-means (never mean-of-ratios):
\[
\Rint = \frac{\operatorname{mean}_x\,\mathrm{JSD}\!\left(p^{\alpha\beta}_x, p^{\mathrm{add}}_x\right)}{\operatorname{mean}_x\,\mathrm{JSD}\!\left(p^{\alpha\beta}_x, p^{00}_x\right)},
\qquad
\Dint^{(q)}_\ell = \Rint_{\mathrm{original}} - \Rint^{(q)}_{\mathrm{patched},\ell},
\]
with the denominator of $\Rint_{\mathrm{patched}}$ \emph{frozen to the unpatched main effect}---a preregistered choice: the patched model's own main effect also moves, and a self-normalized ratio would confound two movements in one number. Unnormalized-JSD shrinkage is a clearly-separated descriptive secondary. $\Rint$/$\Dint$ are formed once from the concatenated prompt set (micro-batch averaging of $\Dint$ is forbidden by the freeze).

\paragraph{Aggregation order and statistics.}
Track A computes every hidden estimand on the same norm-matched $7{\times}7$ composition grid's 36 interior points, prompts, seeds, and aggregation order as the companion study; KT2 uses the frozen subset $\{(1,1),(0.6,1),(1,0.6)\}$ and five frozen boundaries $\{3,8,14,20,25\}$. Seeds are the preregistered outer unit (3/3 direction agreement required); within seed, paired prompt bootstrap with 5{,}000 draws (frozen RNG seed 20260712) yields the CIs reported for KT1 contrasts.

\section{Preregistration timeline and the fourteen predictions}
\label{app:prereg}

\paragraph{Freeze chain.}
The design document was versioned v1.0--v1.6; every version's full text is retained verbatim, and any change to hypotheses, gates, units, doses, statistics, or the decision tree produced a new version rather than a silent overwrite. Table~\ref{tab:freeze} lists the chain. Versions v1.0--v1.4 all predate any hidden-state result. v1.5 is explicitly a \emph{post-KT1} revision---it does not rewrite the KT1 verdict (archived under v1.4 rules) but freezes the hypotheses and pass criteria for the \emph{next} experiments (KT2, direction features, CTL comparison, adversarial arm) before any of their data existed, and adds the ``stable reversal'' rule the original decision tree lacked. v1.6 adds the bf16 dtype audit with its reading rules frozen before the audit ran. A separate point-prediction memo (P1--P6) was frozen after two pilot cells but before the 1{,}296-cell sweep launched.

\begin{table}[h]
\centering\small
\begin{tabular}{lp{10.2cm}}
\toprule
Version & Content of the change \\
\midrule
v1.0 & Initial freeze (2026-08-02): hypotheses, estimators, units, KT1--KT3 gates, decision tree. \\
v1.1 & $\Dint$ denominator fixed to unpatched main effect; determinism null split from ULP floor; patch-quantization audit added. Pre-data. \\
v1.2 & Hardware lock (single A100, \texttt{sdpa}, deterministic, TF32 off); precision lock (bf16 KT1/KT3, fp32 KT2). Pre-data. \\
v1.3 & Batching lock (micro-batch 8, global pad width); aggregation order; equivalence-audit requirement. Pre-data. \\
v1.4 & Per-stage execution manifests (restores the frozen staged order; relaxes no gate). Pre-data. \\
v1.5 & Post-KT1: PK2-1/2/3 frozen before KT2 implementation; F1--F3 and CTL-table analysis preregistration; adversarial arm; stable-reversal rule. \\
v1.6 & bf16 $\G$ dtype audit with frozen reading rules and prior (55\% ``$\G$ real''). \\
P1--P6 memo & Point predictions with priors, frozen before sweep launch. \\
\bottomrule
\end{tabular}
\caption{The preregistration freeze chain. Full documents and sidecars are released with the code.}
\label{tab:freeze}
\end{table}

\paragraph{The fourteen predictions.}
Table~\ref{tab:predictions} lists the original study's fourteen frozen predictions with their prior confidences and outcomes: 7 held, 4 failed, 3 split. The 81 gates cited in the abstract count every preregistered prediction across this document and the follow-up preregistrations (E12--E26), each frozen with its own hash sidecar and judged in a released verdict document. The additionally frozen dtype-audit prior (55\% that bf16 $\G$ was real signal) also fell on the wrong side (\S\ref{sec:bf16}); it is listed separately as an audit belief rather than an experimental prediction.

\begin{table}[h]
\centering\footnotesize
\setlength{\tabcolsep}{4pt}
\begin{tabular}{llp{5.2cm}lp{3.4cm}}
\toprule
\# & Prediction (frozen) & Statement & Prior & Outcome \\
\midrule
1 & P1-direction & $C_{\text{pair}}>0$, both strata, 3/3 seeds & 70\% & \textbf{held} \\
2 & P1-magnitude & AUC ratio in 1.3--2.0 & 70\% & failed (1.02--1.15) \\
3 & P2 & math-stratum contrast $<\!\tfrac12$ of code, still positive & high & \textbf{held} \\
4 & P3 & raw $\G >$ wrapped $\G$, ratio $>1.5$ & 60\% & failed (\emph{reversed}, 3/3) \\
5 & P4 & generation share 4--8\% & high & split (carried-dominance holds; window wrong: mean 12.6\%) \\
6 & P5 & $\cos(\G,B)$ depth profile & high & split (shape yes; variance far larger) \\
7 & P6 & all numerical gates green & high & \textbf{held} (1{,}296/1{,}296) \\
8 & PK2-1 & erasure direction-specific vs.\ all controls & 65\% & \textbf{held} (3/3) \\
9 & PK2-2 & expression gating: $\Dint_{\text{raw}} \gg \Dint_{\text{wrapped}}$ & 55\% & \textbf{held} (3/3, 13$\times$) \\
10 & PK2-3 & pair contrast carried by direction, ratio $>1.3$ & 50\% & \textbf{held} (14$\times$--337$\times$) \\
11 & CTL separation & no CTL/TSV metric separates pair or format ordering & 70\% & split (cumulative $\lVert \I\rVert$ \emph{does} order pairs; format still invisible) \\
12 & F1--F3 & $\ge$1 direction feature passes held-out gate & 45\% & failed (0/3; F1 magnitude-contaminated) \\
13 & Adversarial band & extreme-geometry pairs stay in $\G$ band & 60\% & \textbf{held} \\
14 & Off-diagonal mass & edge stencil $|B| > 2\times$ diagonal & 55\% & failed (max 1.37) \\
\bottomrule
\end{tabular}
\caption{All frozen predictions and outcomes. ``Split'': qualitative claim held, quantitative window failed. No outcome was rewritten after data; reversals are reported as falsifications with the reversed reading tested by \emph{new} frozen predictions (rows 8--10).}
\label{tab:predictions}
\end{table}

\section{Numerical validity audits}
\label{app:numerics}
The four-corner difference is a catastrophic cancellation: the four paths differ at $O(\alpha)$ while $\I_\ell$ is $O(\alpha\beta)$. Every estimator therefore carries its own validity chain, all items frozen before confirmatory execution.

\paragraph{Determinism null (must be exactly zero).}
Pointing all four weight states at the same model and running four independent forwards must produce a \emph{bitwise-zero} four-corner difference under deterministic kernels---this validates determinism only and is explicitly \emph{not} a noise floor (under a deterministic backend it is identically zero and can prove nothing about resolution). All 1{,}296 confirmatory cells passed with exact zeros.

\paragraph{Propagated-ULP cancellation floor (the real noise floor).}
The resolution limit is set by storage precision: we perturb each of the four stored hidden states by $\pm\tfrac12$\,ULP representation uncertainty and re-form the four-corner difference, reporting per-layer floor RMS (mean/p99/max) under frozen draws and RNG seed. One implementation subtlety is instructive: propagating the jitter in float32 \emph{silently returns zero}---sub-ULP perturbations round back to the original value 100\% of the time---so the propagation is done in float64 and validated by the exact $2^{15}$ ratio between fp32 (23 mantissa bits) and bf16 (8 mantissa bits) floors. At the main composition point $(1,1)$ the signal-to-floor ratio is min 10.2 / median 38.2 / max 92.7 across confirmatory layers; layers below ratio 1 would be flagged \emph{below numerical resolution} and barred from mechanistic interpretation (none were). Boundary 0 (embedding) is excluded as a definitional zero, not a resolution failure.

\paragraph{First-block identities.}
With LoRA leaving the embedding untouched, $\I_1 = 0$ exactly and $\I_2 = \G_1$ to within the ULP floor. Both identities are asserted per cell; they jointly validate the hook boundary, the reference-state construction, and the block-map definition.

\paragraph{Bitwise equivalence audit of performance refactors.}
Single-GPU feasibility required three optimizations---capture-and-replay direct block invocation, prompt micro-batching, and (initially) \texttt{logits\_to\_keep=1}. Each had to prove semantic invariance \emph{on the locked A100} (kernel behavior is part of the estimand; equivalence does not transfer across hardware). The audit caught two real issues before any confirmatory data existed. (i) \texttt{logits\_to\_keep=1} changes the lm-head GEMM shape and hence accumulation order, producing a 1-ULP deviation (max $0.0625$ at $|$logit$| \in [16,32)$) against the full-sequence path; the optimization was removed rather than tolerated. (ii) \texttt{output\_hidden\_states} returns the \emph{final-norm-transformed} state at the last boundary, not the last block's raw output---mixing spaces in the ledger (observed direct-block discrepancy 67.94). Fixed with a forward hook capturing the raw output, a uniform \texttt{residual\_state()} accessor, and a regression test locking the contract. After both fixes, all equivalence checks are exactly $0.0$ at every audited depth.

\paragraph{The bf16 dtype audit (frozen reading rules).}
Recomputing $\G_\ell$ on the identical $\bar h_\ell$ with block weights and arithmetic cast to float32: fp32/bf16 RMS ratio median 0.230 (\csafety) and 0.136 (\cmath), minima 0.073/0.040; direction cosines median 0.239/0.152, minimum $-0.031$. Both frozen reading rules place this in the ``$\G$ is predominantly weight-rounding artifact'' branch (\S\ref{sec:bf16}). A related earlier catch: an all-bf16 mixed-derivative stencil---which extracts an $O(\epsilon^2)$ signal from $O(\epsilon)$ differences, a worse cancellation than $\G$ itself---produced pure noise ($\approx$50$\times$ the true value), exposed by its own frozen gates (probe convergence $\approx 1.0$ everywhere; $|B| > 50|\G|$ is physically impossible). All stencils were moved to fp32 block evaluation with their own floors. The general lesson we would offer practitioners: \emph{every new activation-difference estimator needs its own propagated noise floor and convergence check, or it will discover noise}.

\section{Control constructions and the cross-prompt derangement}
\label{app:controls}
At each frozen boundary $\ell \in \{3,8,14,20,25\}$ the intervention replaces the merged model's residual state $h^{AB}_\ell$ with $h^{AB}_\ell + d^{(q)}_\ell$ and continues the forward pass in float32. The correct erasure has displacement $d^{(\mathrm{correct})} = -\lambda \I_\ell$ per prompt. Every control is \emph{row-wise norm-matched}: its raw displacement is rescaled per prompt to $\lVert \I_\ell(x) \rVert$, so all conditions apply identically-sized perturbations and differ only in structure.

\begin{itemize}\itemsep2pt
\item \textbf{Identity/no-op} ($\lambda=0$): validates that the hook mechanism itself leaves the output bitwise unchanged.
\item \textbf{Wrong-pair}: the additive reconstruction built with the \emph{other} task pair's second axis, $d = (h^{A0} + h^{0B'} - h^{00}) - h^{AB}$, where $B'$ is a third adapter (for \csafety{} the math axis and vice versa). Preserves the additive construction; destroys task-pair identity.
\item \textbf{Coefficient-mismatch}: the frozen wrong-coefficient template $(+1,-1,+1)$ in the same affine subspace, $d = (h^{A0} - h^{0B} + h^{00}) - h^{AB}$, norm-matched. Same subspace, same prompts, wrong combination.
\item \textbf{Cross-prompt}: the correct additive displacement computed on a \emph{donor} prompt $x'$ and applied to $x$. Donors come from a deterministic length-binned derangement frozen before implementation (master seed 20260802): prompts are binned by token length, and within each bin a derangement guarantees no prompt donates to itself while keeping donor and recipient lengths comparable. This is the direct null for prompt-conditioned structure and the second arm of the dose ladder.
\item \textbf{Distance-matched random}: per-prompt Gaussian direction scaled to $\lVert \I_\ell \rVert$; frozen as a secondary sensitivity control, never a substitute for the structured nulls.
\end{itemize}

\paragraph{Overlap accounting.}
For every condition we report $\cos(d^{(q)}, -\I_\ell)$ and interpret specificity contrasts conditional on it (frozen rule \S8.3/\S8.8 of the design). The controls span overlap cleanly: correct $=1.0$ by construction; wrong-pair $\approx 0.25$ (its additive scaffold shares the $A$-axis); coefficient-mismatch, cross-prompt, and random near zero. The observed effect is approximately proportional to the projection on $-\I_\ell$ (at boundary 14: correct $\cos = 1.00$, $\Dint = +0.028$; wrong-pair $\cos = 0.25$, $\Dint = +0.007$; its depth-averaged $\Dint$ is nonetheless negative)---which is exactly why the paper claims \emph{cross-term-direction specificity} rather than task-identity specificity beyond the direction (\S\ref{sec:conclusion}).

\paragraph{Patch-quantization audit.}
Writing patched states back to bf16 could silently cancel low-dose rungs ($\lVert \mathrm{bf16}(h - \lambda\I) - (h - \lambda\I)\rVert / \lVert \lambda\I \rVert$ exceeding a frozen 0.3 threshold would mark a rung \emph{not actually applied}). The v1.2 freeze resolved this preemptively by running all of KT2 in float32 (with TF32 off, since its 10-bit mantissa would defeat the purpose); the audit is retained as residual-risk monitoring and never triggered.

\section{Amplification profiles (exploratory)}
\label{app:amplification}
Everything in this appendix is computed from existing immutable cell records after the confirmatory verdicts; it motivated no gate and is labeled exploratory throughout.

\paragraph{The exclusion arithmetic (confirmatory input to \S\ref{sec:amplify}).}
At the $(1,1)$ code cell, if layers contributed independent increments of the measured sizes $\lVert \G_\ell \rVert$, cumulative growth would be bounded by $\sqrt{\sum_\ell \lVert \G_\ell \rVert^2} = 0.088$ (incoherent) and by $\sum_\ell \lVert \G_\ell \rVert = 0.369$ even under perfect alignment of every increment. Observed: $\lVert \I_{28} \rVert = 0.974$---11$\times$ and 2.6$\times$ the respective bounds, implying an average multiplicative gain of $\approx 1.19$ per block on the carried term. Since the bounds are arithmetic in the measured quantities, no per-layer injection account survives regardless of the increments' provenance (including the rounding-artifact component of bf16 $\G$, which only strengthens the exclusion by shrinking the true injections).

\paragraph{Layer-resolved gain.}
The per-block gain $\lVert \I_{\ell+1} \rVert / \lVert \I_\ell \rVert$ is above 1 through most of the depth, with pair- and format-dependent profile: gains are similar across the two task pairs early and diverge in the second half of the network---consistent with \S\ref{sec:metrics}, where \emph{cumulative} magnitude (post-amplification) coarsely orders pairs even though local generation does not. Across the 15-pair capture grid, cumulative $\lVert \I \rVert$ spans a 3$\times$ range (0.084--0.256) on a generation floor that is nearly pair-independent.

\paragraph{Super-coherent amplification by format.}
Defining the super-coherence ratio as observed $\lVert \I \rVert$ over the coherent accumulation bound: raw prompts 1.0--1.7$\times$, instruction-wrapped prompts 3.8--4.1$\times$---amplification is \emph{strongest} exactly where expressed interference is smallest.

\paragraph{Depth relocation (3/3 seeds).}
\label{par:relocation}
Splitting mean gain at normalized depth 0.5: wrapped early-gain 1.55 vs.\ raw 1.36, while the raw$-$wrapped \emph{late}-gain contrast is $+0.019/+0.012/+0.031$ across seeds---positive 3/3. The wrapper front-loads amplification and suppresses it late; raw amplification persists to the output. This profile was recorded (as an exploratory expectation, not a gate) \emph{before} KT2 ran, where it correctly anticipated the causal gating result: erasure finds interference to remove on raw prompts and almost none under the wrapper (\S\ref{sec:gating}).

\section{Second-family replication (Llama-3.2-1B)}
\label{app:replication}
A reduced replication with the base family as the \emph{only} varied factor: \texttt{meta-llama/Llama-3.2-1B} (commit-pinned; 16 blocks, GQA, 128k tokenizer), nine LoRA adapters trained with the identical recipe (datasets, samples, epochs, LoRA shape, seeds), intervention boundaries $\{2,5,8,11,14\}$ mapped from the frozen Qwen set at equal normalized depths, and a new 9-delta core-median norm radius recorded before any kill-test cell ran. Four predictions (RP1--RP4, with priors 70/75/60/high\%) were frozen before any adapter on this family existed. Execution: 54 reduced KT1 cells (KT2 subgrid $\times$ \{controlled code, raw, wrapped\} $\times$ 3 seeds $\times$ 2 pairs), the amplification-bound analysis, and the full 54-run KT2 with per-prompt persistence, under the same numerical-validity chain---which passed unmodified on the new architecture (RP4: bitwise determinism nulls, first-block identities, zero equivalence audits).

\paragraph{RP1 (band + output mismatch): held, with reduced margins.}
The bf16 $\G$-AUC band ratio across all 18 $(1,1)$ cells is 1.201 (gate $<$1.3; absolute level $\approx$0.0013, an order below Qwen's---hidden width and norm scale differ, and only within-band uniformity is the cross-family claim). Output interference again exceeds what the band allows: pair ratio 1.51 (gate $>$1.5; per-seed 1.59/1.82/1.06), format ratio 2.03 (gate $>$2; per-seed 2.82/1.71/1.93). Both gates pass at the frozen mean-level reading but are seed-variable and far smaller than Qwen's 2$\times$/20$\times$; we grade this held-with-reduced-margins. Not lost in the reduction: $\G_{\text{wrapped}} > \G_{\text{raw}}$ replicates 3/3 seeds---the falsified-then-reversed KT1 prediction is not a Qwen artifact.

\paragraph{RP2 (injection exclusion): held, more strongly than Qwen.}
Cumulative-to-bound ratios per condition: 6.7--24.9$\times$ incoherent (gate $>$3), 2.19--7.85$\times$ coherent (gate $>$1.2). The depth-relocation profile of Appendix~\ref{app:amplification} also transfers: super-coherent amplification is again largest under the wrapper (5.4--7.9$\times$ vs.\ 2.2--3.2$\times$ raw), and the raw$-$wrapped late-gain contrast is positive 3/3 seeds ($+0.042/+0.053/+0.070$).

\paragraph{RP3 (causal triple): held; the gating leg by its frozen letter only.}
With the frozen bootstrap template: (a) $\Dint_{\text{correct}} = +0.0233/+0.0333/+0.0192$, all 15 intervals (effect and four specificity contrasts per seed) exclude zero. (c) Implied pair ratios $\Dint(\text{\csafety})/\Dint(\text{\cmath}) = 2.84/5.46/2.56$ (gate $>$1.3, 3/3), contrast CIs excluding zero---and for the seed whose \emph{output} pair ratio is only 1.06, erasure still separates the pairs 2.6$\times$: the direction-carried contrast is more stable than the output ratio itself. (b) $\Dint_{\text{raw}} - \Dint_{\text{wrapped}} > 0$ holds 3/3 by sign, satisfying the frozen criterion, but margins are weak ($+0.0052^{*}/+0.0019/+0.0039^{*}$; the middle CI crosses zero) and the removable-interference ratio is only $\approx$2$\times$---consistent with this family's weak output-side format collapse. The gate's \emph{direction} transfers; its \emph{strength} is a family property.

\section{Regrowth after erasure (preregistered follow-up)}
\label{app:regrowth}
Three predictions were frozen before any regrowth data existed (priors 60/60/65\%). Setup: at each frozen boundary $b \in \{3,8,14,20\}$, apply the $\lambda{=}1$ correct erasure and continue the forward pass; at every later frozen boundary (plus the raw final-block output) form the post-patch cross-term against the \emph{unpatched} marginals, $\I^{\mathrm{post}}_\ell = \tilde h^{AB}_\ell - h^{A0}_\ell - h^{0B}_\ell + h^{00}_\ell$, and report the per-prompt recovery ratio $\rho_\ell$ and direction cosine to $\I^{\mathrm{orig}}_\ell$. Scope: 18 cells (2 pairs $\times$ 3 strata $\times$ 3 seeds, $(1,1)$, float32).

\paragraph{Outcomes.}
(PE1-1, \emph{falsified}): final-boundary $\rho$ after erasing at depth 8 was predicted in $(0.3, 0.95)$; observed $0.987/0.998/0.992$ across seeds---regrowth is essentially complete, outside the window on the ``transient'' side. (PE1-2, \emph{held}): direction memory $\cos(\I^{\mathrm{post}}, \I^{\mathrm{orig}}) > 0.5$ predicted; observed $0.987/0.987/0.986$. (PE1-3, fail by letter): $\rho$ decreasing in patch depth holds in 2/3 seeds; the third inverts between depths 3 and 8 by $+0.0003$, a noise-level violation we report as the frozen criterion requires. Full profile (seed means): $\rho = 1.000/0.992/0.944/0.843$ erasing at depths $3/8/14/20$.

\paragraph{Interlock with the causal effect.}
Decomposing the KT2 effect by patch depth from the per-prompt records (\csafety, code stratum, mean over grid): $\Dint \approx 0$ at depths 3--8 ($+0.0005/-0.0012$, $+0.0009/+0.0037$, $+0.0001/-0.0076$ per seed), rising to $+0.0374/+0.0408/+0.0260$ at depth 20 and $+0.0545/+0.0582/+0.0366$ at depth 25. The unrecovered share $1-\rho$ and the durable output effect $\Dint$ rise together: what erasure removes is what the remaining depth cannot rebuild. The regeneration reading was subsequently subjected to a preregistered basin test (gate frozen first: final-boundary $\cos > 0.8$ from every start): patching at depths 8 and 14 with correct erasure at $\lambda \in \{1,2\}$ and all four norm-matched structural controls, the downstream cross-term reconverges to the original direction from \emph{every} start---per-condition means $0.90$--$0.99$, worst single cell $0.83$, 18/18 (both pairs, 3/3 seeds). Convergence of direction is compatible with the divergent \emph{output} effects of \S\ref{sec:erasure}: wrong-direction patches commit their damage to the expression pathway before late-network regrowth completes, exactly as the depth decomposition shows.

\paragraph{Relation to emergent self-repair.} The Hydra effect \citep{mcgrath2023hydra} and its systematic follow-up \citep{rushing2024selfrepair} show that ablating a transformer component is partially compensated downstream: an output-side, incomplete, prompt-noisy recovery of the component's direct effect, with final-LayerNorm rescaling explaining at most 30\%. The regeneration here is a different object: what returns is a four-path \emph{differential} coordinate, rebuilt to 99\% of its norm at cosine 0.99 from three marginal paths the intervention never touched, while the state itself never rejoins its trajectory and no direction-specific restoring force exists (Appendix~\ref{app:recovery}). LayerNorm rescaling acts within the patched trajectory and could in principle contribute to the recovered \emph{magnitude}; it does not by itself supply the direction memory the basin test measures from every start, nor the marginal-path provenance the ledger decomposition establishes. A quantitative accounting of how much of the regrowth known self-repair mechanisms explain is open.

\section{SurgeryV2 representation bias, head-to-head}
\label{app:surgerybias}
Two predictions frozen before any bias data existed (priors 70/50\%). Operationalization: in the factorial setup the task experts are the marginal paths, so layerwise bias is $\lVert h^{AB}_\ell - h^{A0}_\ell \rVert$ (and against the right expert, $\lVert h^{AB}_\ell - h^{0B}_\ell \rVert$), last prompt token, bfloat16 measurement lock, full-depth AUC over the frozen boundaries; 36 cells (2 pairs $\times$ 6 strata $\times$ 3 seeds, $(1,1)$).

\paragraph{Format axis (frozen prediction: wrapped $\ge$ raw, all 6 pair$\times$seed conditions).}
Held in 5/6; one condition (\csafety, seed 42) inverts by $0.8\%$ (16.36 raw vs.\ 16.23 wrapped)---reported as a falsification of the letter. The substantive comparison is one-sided everywhere: bias ratios wrapped/raw span $0.99$--$1.34$ while the expressed-interference ratio on the same prompts is $\sim$0.05 (a 20$\times$ collapse). Representation bias does not track expression in any condition.

\paragraph{Pair axis (frozen two-sided).}
Bias orders the pairs as the output does: bias(\csafety)/bias(\cmath) $= 2.83/3.05/2.96$ across seeds on code prompts. This places representation bias in the same family as cumulative $\lVert \I \rVert$ (\S\ref{sec:metrics}): a coarse pair-level magnitude signal---expected, since it contains the amplified cumulative term---with no visibility into expression gating.

\section{Intervention validity: introduced divergence and context sensitivity}
\label{app:validity}
Causal interventions can displace states off the model's natural distribution, where they recruit pathways that never fire naturally; \citet{grant2026divergent} accordingly ask intervention papers to (i) report the representational divergence they introduce and (ii) test interventions for context sensitivity. We address both.

\paragraph{The displacement is the estimand's own counterfactual, not an arbitrary perturbation.} Our patch sends $h^{AB}_\ell \mapsto h^{AB}_\ell - \lambda \I_\ell$, so at $\lambda{=}1$ the patched state is exactly $\bar h_\ell = h^{A0}_\ell + h^{0B}_\ell - h^{00}_\ell$: the additive reconstruction against which expressed interference is defined, and the state the network would occupy if composition were additive at that depth. This is the minimal displacement that answers the causal question, in contrast to the multiplicative feature scalings (up to $15\times$) that motivate the concern.

\paragraph{Introduced divergence, measured.} We report the displacement itself, not only its behavioral consequences. Over the same six cells (2 pairs $\times$ 3 seeds, code prompts, float32), the $\lambda{=}1$ displacement is $\lVert \I_\ell \rVert / \lVert h^{AB}_\ell \rVert = 1.5$--$3.6\%$ of the state norm at depth 3, rising to $5.3$--$14.3\%$ at depth 25; measured against the natural prompt-to-prompt spread of merged states at the same depth, it is $0.06$--$0.13$ of that spread early and at most $0.60$ late. The patched state therefore stays inside the envelope of variation the model already exhibits across prompts---a different regime from the multiplicative feature scalings that motivate the concern. Output-side the divergence is correspondingly small: patched-versus-unpatched first-token JSD $0.003$ against a main effect of $0.236$, top-1 agreement $0.94$--$0.99$, and 92--99\% of continuation tokens preserved (\S\ref{sec:erasure}). Behavioral invariance alone cannot exclude dormant divergences---precisely \citeauthor{grant2026divergent}'s point---which is why we also report the geometry above, and treat $\lambda > 1$ as an explicitly off-manifold regime where only ordered contrasts, never absolute effects, are interpreted.

\paragraph{The displacement's own depth profile does not explain the effect.} Relative displacement grows with depth, so one might ask whether the depth profile of $\Dint$ merely tracks how much we move the state. It does not, quantitatively: from depth 3 to 25 the displacement grows $3$--$4\times$ while $\Dint$ grows roughly $100\times$ ($+0.0005 \to +0.055$). The erasure is complete at every depth by construction ($\lambda{=}1$ removes the entire cross-term present there), and what differs is whether propagation rebuilds it---measured directly as $\rho$ (Appendix~\ref{app:regrowth}), not inferred. The angle sweep isolates the same point with displacement norm held fixed.

\paragraph{Context sensitivity.} A hidden-pathway artifact would show up as an effect that exists in one configuration and vanishes or reverses arbitrarily elsewhere. The same intervention is applied at five depths, on three prompt distributions, for two task pairs, at three model points, and---in the angle sweep (Appendix~\ref{app:orthogonal})---at five orientations with depth and norm held fixed. The outcome is not configuration-idiosyncratic but systematic in the intervention's geometry: $\approx 0$ at early depths and rising with $1-\rho$; monotone in $\cos(\delta, -\I_\ell)$; sign-determined by that angle rather than by displacement size. A dormant-pathway account would have to reproduce this two-parameter structure across all four axes, which is a far stronger commitment than the effect itself.

\section{Auditing merging methods with the ledger}
\label{app:merge}
Four weight-space merging methods, applied post hoc to the same LoRA task vectors with no training: \textbf{task arithmetic} (the composition used throughout this paper), \textbf{trim-only} (keep the top 20\% of $|\Delta|$ per task), \textbf{TIES} \citep{yadav2023ties} (trim, sign-elect, disjoint mean) and \textbf{DARE} \citep{yu2024dare} ($p{=}0.9$ random drop with $1/(1{-}p)$ rescaling). For method $M$ the merged weights are $W_0 + \alpha\sum_t \Delta_t^M$ and the additive counterfactual is built from that method's \emph{own} marginals $W_0 + \alpha\Delta_t^M$, so $\I_\ell$ always measures the non-additivity of the composition step and is comparable across methods; deltas are written into the base weights with the adapter path disabled and restored exactly afterwards. Grid: 3 seeds $\times$ 4 methods $\times$ \{$k{=}2,\alpha{=}1$; $k{=}6,\alpha{=}1$; $k{=}6,\alpha{=}1.5$\}, scored in the templated regime. Pipeline check (not a gate): task arithmetic at $k{=}2$ gives $\Rint = 0.0031/0.0044/0.0025$ against $0.0031/0.0044/0.0025$ measured independently in Appendix~\ref{app:degradation}---the weight-writing and adapter-scaling paths agree to three or four significant figures.

\paragraph{Result.} In the collapse cell ($k{=}6$, $\alpha{=}1.5$), seed-mean $\Rint$ and code score: TIES $0.93 / 0.84$; trim-only $0.77 / 0.73$; DARE $0.67 / 0.18$; task arithmetic $0.64 / 0.42$. The frozen prediction that some method would cut $\Rint$ by $\ge$30\% failed 0/3. The frozen pooled-correlation prediction ($\rho \le -0.4$) passed at $-0.42$, but the within-cell correlations are $+0.20$ ($k{=}2$), $+0.07$ ($k{=}6$) and $+0.93$ ($k{=}6$, $\alpha{=}1.5$), so the pooled value reflects cell composition, not method ranking. We report both.

\paragraph{A common counterfactual (preregistered follow-up).} The per-method convention above leaves TIES' marginal semantically different from a standalone expert (its limit below), so a frozen follow-up recomputed $\Rint$ for all four methods against ONE shared reference: the additive prediction from the task-arithmetic marginals ($W_0 + \alpha\Delta_t$, the actual standalone experts) and the same base logits. The pipeline check is exact: $\Rint^{\text{common}}$(task arithmetic) reproduces the frozen per-method value to 10 decimal places, 3/3 seeds. The frozen expectation (55\%: still no correct ranking) held, in a stronger form than predicted: under the common reference the output-side ratio carries almost no method information at all---four-method range $0.006/0.015/0.006$ against behavioural spreads of $0.6$--$0.9$, Spearman $0.0/-0.4/+0.4$. The between-method differences of the per-method convention ($0.64$--$0.93$) are therefore largely a property of the counterfactual choice, not of the merged models: under either convention the output-side ratio cannot rank methods, while the state-space measures rank the same models correctly (Appendix~\ref{app:merge7b}).

\paragraph{Reading, and its limits.} The two methods that preserve behaviour both \emph{trim}; the two that collapse both keep full-support displacements (DARE drops 90\% of entries but rescales to preserve norm). What these methods control is the support and magnitude of the displacement; $\Rint$ measures non-additivity, and in the collapse regime the two move in opposite directions. This reading is post hoc and untested. Four limits belong with it: TIES and DARE were designed for dense full-model deltas, and their statistical assumptions may not transfer to rank-16 LoRA deltas---the audit's method axis is scoped accordingly; for TIES the per-task marginal is not a standalone expert (it is that task's share of the elected merge, divided by the agreement count), which changes what its additive counterfactual means; four methods give a coarse rank correlation, so the claim rests on 3/3 seed agreement in direction; and the behavioural instruments see collapse but not competence (Appendix~\ref{app:degradation}), which is why the audit is stated for the collapse cell only.

\section{The metric family on the same models, and the audit at 7B}
\label{app:merge7b}
Two preregistered follow-ups scope the merging audit (Appendix~\ref{app:merge}); all gates below were frozen before any of their data existed.

\paragraph{The family, same models, same scores.} On the identical merged models and prompts, with behavioural scores read from Appendix~\ref{app:merge}'s immutable records (nothing regenerated), we computed four family members: our output-side $\Rint$; SurgeryV2-style representation bias ($\mathrm{mean}_\ell\,\mathrm{mean}_x \lVert h^{\mathrm{merged}}_\ell - h^{\mathrm{expert}}_\ell\rVert$); a CTL-style state residual ($\mathrm{mean}_\ell\,\lVert h^{AB}_\ell - \bar h_\ell\rVert / \lVert h^{AB}_\ell - h^{00}_\ell\rVert$---the same relative non-additivity as $\Rint$, read before the readout); and mean pairwise parameter cosine. Two frozen expectations both \emph{failed}, in the direction the preregistration marked as the more valuable outcome: some members do rank correctly (65\% prior against), and $\Rint$ opposes rather than tracks bias (55\% prior for agreement). Collapse-cell seed means:

\begin{center}\small
\begin{tabular}{lcccc}
\toprule
 & code score & $\Rint$ & bias & CTL residual \\
\midrule
TIES & \textbf{0.84} & 0.93 & \textbf{26.7} & \textbf{2.00} \\
trim-only & 0.73 & 0.77 & 31.2 & 2.22 \\
task arithmetic & 0.42 & 0.64 & 38.1 & 2.33 \\
DARE & 0.18 & 0.68 & 44.7 & 2.37 \\
\bottomrule
\end{tabular}
\end{center}

\noindent Bias and the residual align with score rank-for-rank (Spearman $-0.8$ to $-1.0$, 3/3 seeds); $\Rint$ is their mirror image ($+0.8$ to $+1.0$); parameter cosine is weakly backwards ($+0.4$ to $+0.8$).

\paragraph{The audit at 7B with real endpoints.} Qwen2.5-7B carries three adapters (math, code, safety), so collapse is approached through the coefficient: 4 methods $\times$ $\alpha \in \{1, 1.5, 2\}$ $\times$ 3 seeds, templated regime, endpoints fixed before any data: GSM8K \emph{test} split, first 200 problems, exact final-number match \citep{cobbe2021gsm8k} (training used the train split), and HumanEval, 164 problems with executed unit tests \citep{chen2021codex}; templated mode is primary with completion mode as instrument control, the code endpoint taking the better of the two. The instrument check passed \emph{before} measurement for the first time in four behavioural attempts: base scores 0.315 (GSM8K) and 0.701 (HumanEval), and merged models at $\alpha{=}1$ \emph{exceed} base on GSM8K (up to 0.60)---these instruments see competence, not only collapse.

\paragraph{Collapse by the letter, and the gates.} The frozen collapse criterion (mean code-endpoint drop $\ge$25\% against each method's own expert, at the highest such $\alpha$) fires in one seed of three ($\alpha{=}2$: 30.2\%, against 24.7\% and 23.2\% in the others---a binary gate missing by 0.3 points, reported as such). In that cell the state-space measures rank the four methods correctly against \emph{both} real endpoints (Spearman $-1.0$; GSM8K and HumanEval give identical method orderings: TIES 0.67/0.41, trim 0.61/0.31, task arithmetic 0.52/0.13, DARE 0.31/0.06), and trimming wins as at 1.5B (frozen prediction held). Two frozen expectations \emph{failed}: the output-side reversal did not persist (Spearman($\Rint$, score) $= -0.2$, gate required $\ge 0$), and $\Rint$ no longer opposes bias ($+0.2$, gate required $\le 0$). Across the full nine-cell grid the 1.5B dissociation is absent everywhere---$\rho(\Rint, \mathrm{bias}) = +0.2$ to $+1.0$---and $\Rint$ ranks the methods correctly in seven of nine cells, but its four-method range collapses to $0.06$ in the deepest-degradation cell while the state-space measures stay at $-1.0$.

\paragraph{The three-layer decomposition (preregistered reanalysis).} A review asked whether ``magnitude points the wrong way'' is partly a normalization artifact. On the frozen payloads only (statistics never previously computed; gates frozen before computing them), we correlated the \emph{absolute} interaction JSD, the normalized $\Rint$, and the main-effect denominator with behavioural degradation, per axis. On the \emph{coefficient} axis (E14's $k{=}6$ column and the 7B per-method sweeps; 15 instances) the absolute numerator ranks degradation no worse than $\Rint$ in 11/15---the normalization critique partly holds there. On the \emph{method} axis it does not: under the common counterfactual the absolute interaction is as method-blind as the ratio (relative range under 10\% of the behavioural range, 3/3 seeds). And on the \emph{format} axis the failure lives in the denominator (absolute $0.4$--$1.0\times$, ratio $13\times$, behaviour $20\times$). Three axes, three distinct failure modes; the blanket claim is retired in favour of this table (\texttt{three\_layer\_verdict.json}).

\paragraph{Reading, and limits.} The reversal of Appendix~\ref{app:merge} is thereby bounded to the small-scale collapse regime, and the recommendation the two experiments jointly license---rank merging methods by state-space deviation---holds at both scales and survives its preregistered gate. Consistent with \S\ref{sec:theory}, the failure lives in the state-to-output readout, whose gain is regime- and scale-dependent: backwards where the 1.5B collapse readout compresses the cross-term direction, signal-free where deep degradation saturates numerator and denominator together. Limits: four methods give coarse rank correlations; the 7B collapse cell exists in one seed by the letter; collapse at 7B is approached through $\alpha$ rather than task count, so the two collapse cells are not the same parameterization; and the HumanEval mode comparison drifts across cells (the endpoint takes the better mode, per the frozen instrument clause).

\section{The degradation map: where merging actually breaks}
\label{app:degradation}
Behavioural validation failed three times (Appendix~\ref{app:behavior}). Before designing a fourth attempt we asked the prior question---\emph{does a large, seed-stable degradation exist anywhere accessible?}---by measuring, with no intervention, a grid over the number of merged tasks $k \in \{1,2,4,6\}$ and a uniform coefficient $\alpha \in \{0.5,1,1.5\}$, scored in the templated regime that matches the training format, with $\Rint$ recorded in the same cells. Two predictions were frozen first; both failed, and the reasons are the content.

\paragraph{The frozen normaliser was degenerate: the experts do not beat the base model on these endpoints.} We had specified degradation as a fraction of the expert's advantage over base. That advantage is non-positive here: on code, base scores $0.958$ against the expert's $0.917$; on math both score $0.875$---while the same adapters move the output distribution enormously (main-effect JSD $\approx 0.56$). The criterion was therefore unsatisfiable and we record it as failed by its letter. This is the third instance of one disease (Appendix~\ref{app:behavior}): \textbf{our behavioural instruments can see collapse but not competence.}

\paragraph{Degradation exists---but not where our causal experiments are.} In absolute terms the merge does collapse at the edge of the grid: at $k{=}6, \alpha{=}1.5$ the code score falls from $0.917/0.917/0.889$ (single-task) to $0.403/0.194/0.667$ (3/3 seeds down, mean $-0.49$), and math falls to $0.250$ in one seed. Monotonicity in $k$ nonetheless failed (3/3): at $\alpha{=}1$ the mean absolute drop is $0.056$ ($k{=}2$), $0.032$ ($k{=}4$), $0.120$ ($k{=}6$), with one seed \emph{improving} at $k{=}4$---interference from different tasks can partially cancel, as the preregistration's stated counter-argument anticipated.

\paragraph{The gate is task-count dependent, which bounds \S\ref{sec:gating}.} In the same templated regime where two-task merging gives $\Rint \approx 0.003$, expressed interference rises steeply with $k$ at $\alpha{=}1$: $0.0033$ ($k{=}2$) $\to 0.198$ ($k{=}4$) $\to 0.821$ ($k{=}6$), a factor of $\sim$250. At $k{=}6$ the interaction JSD is of the same order as the main effect. The instruction template pins the output distribution when two task vectors are added; it does not when six are. \textbf{Our gating result is measured at $k{=}2$ and should be read with that scope.}

\paragraph{$\Rint$ is not a sufficient statistic for degradation.} Across all 27 grid cells the correlation between $\Rint$ and absolute drop is $+0.39$; \emph{within} $k{=}6$ it is $-0.32$, because $\Rint$ saturates near 1 while degradation keeps growing with $\alpha$. Expressed interference tracks the number of merged tasks; degradation tracks both $k$ and $\alpha$. At the seed level the association is visible but the sample is three: at $k{=}4, \alpha{=}1$ the seed with $8\times$ lower $\Rint$ is the only one without degradation.

\paragraph{Why this closes the behavioural question honestly.} Our causal experiments all sit at $k{=}2, \alpha{=}1$, where the merge is nearly lossless on these endpoints; the degradation lives at $k{\ge}6, \alpha{=}1.5$, where interference is also strongly expressed. The two regimes are disjoint. That is a structural reason for the failure of behaviour-level validation---not a matter of instrument precision---and it names the coordinates a future causal experiment should use.

\section{The behavioural loop in the collapse regime}
\label{app:repair}
Both external audits of this work asked the same question: the causal experiments sit at $k{=}2, \alpha{=}1$, where the merge is near-lossless---does erasing the cross-term repair behaviour \emph{where merging actually breaks}? A preregistered fourth behavioural attempt (both outcomes' obligations frozen in advance, prior 35\% for repair) ran the intervention in E14's collapse cell itself.

\paragraph{Design.} $k{=}6$ adapters at $\alpha{=}1.5$, templated regime, float32, 3 seeds. The $k$-way cross-term $\I_b = h^{\text{merged}}_b - \big(h^{00}_b + \sum_t (h^t_b - h^{00}_b)\big)$ is patched at the last prompt position at boundary $b \in \{8, 14, 20, 25\}$ during prefill (the KT2 intervention point; the KV cache carries it into greedy generation). Four norm-matched arms: none, correct ($-\I$), $+\I$, random-orthogonal (the E10/E13 construction and seed). Endpoints: the templated code and math scores---\emph{independent of the additive counterfactual}, with collapse-cell sensitivity certified by E14's frozen payload before any data existed. Pipeline check: the no-patch baselines reproduce E14's frozen collapsed scores bitwise ($0.403/0.194/0.667$), 3/3 seeds.

\paragraph{Outcome: no repair (frozen gate failed 0/3), and an asymmetry.} No late boundary reaches the preregistered $\ge$25\% recovery of the collapse gap (per-boundary means $-0.12$/$-0.87$/$0.00$ at $b{=}14/20/25$). At the final boundary the result is exact: erasing the entire cross-term---a displacement of $1.9$--$2.1\times$ the state norm here---leaves the collapsed code score \emph{bitwise unchanged} in all three seeds, while the equal-norm $+\I$ arm is destructive ($0.40\to0.00$; $0.67\to0.22$). The residue can do damage in the $+$ direction; its removal repairs nothing. The direction-specificity gate was conditional on repair and is recorded as not triggered.

\paragraph{Reported both ways (unregistered directions, no claims).} At $b{=}8$---excluded from the gate because regrowth was expected to undo early erasure---two of three seeds show substantial recovery ($+0.32/+0.31/-0.50$), the reverse of the first-token depth profile; the math endpoint (8 items, 0.125 granularity) shows no 3/3 pattern anywhere; at $b{=}20$ the correct arm is harmful in 2/3 seeds---mid-depth erasure of a displacement this large does damage of its own. In this cell $\lVert\I_b\rVert/\lVert h_b\rVert = 0.83$--$2.36$: the intervention is far outside the small-displacement regime of Appendix~\ref{app:validity}, so only ordered contrasts between equal-norm arms are interpreted, per the paper's standing policy.

\paragraph{The small-dose ladder (preregistered follow-up).} The $\lambda{=}1$ null leaves a cancellation reading: removal of a harmful component and off-manifold damage of the same size. The first-order logic discriminates: a carried component gives an $O(\lambda)$ benefit, off-manifold damage is $O(\lambda^2)$, so the small-dose slope decides. We ran $\lambda \in \{0.1, 0.25, 0.5, 1.0\}$ (at $\lambda{=}0.1$ the displacement is 17--21\% of the state norm, comparable to Appendix~\ref{app:validity}'s regime) at $b \in \{14, 25\}$ with a continuous endpoint fixed in advance: teacher-forced NLL of frozen expert reference continuations, whose dynamic range was certified \emph{before} any arm data (collapse gap $= 34$--$50\times$ the expert's cross-seed spread); the $\lambda{=}1$ discrete scores reproduce the main experiment bitwise. The frozen repair gate (30\% prior) \emph{failed}: at the final boundary the correct arm's dose curve is flat---$\Delta$NLL $\approx -0.002$ regardless of $\lambda$, a tenfold dose change with no effect---while the orthogonal arm degrades quadratically ($-0.073$ to $-0.087$ at $\lambda{=}1$), exactly the bystander signature; at mid-depth a $+0.003$/$+0.004$ small-dose signal appears in 2/3 seeds, is absent in the third, and even taken at face value bounds any carried share at a few percent of the collapse gap. The cancellation reading is excluded across the full dose range.

\paragraph{A second collapse parameterization (preregistered follow-up).} The loop was repeated where collapse is reached through the coefficient instead of the task count: Qwen2.5-7B, three adapters at $\alpha{=}2.0$ (the deepest degradation E18 measured), real-benchmark endpoints, arms and construction unchanged, $b \in \{14, 25\}$. Pipeline check: the no-patch baselines reproduce E18's weight-written $\alpha{=}2$ scores \emph{bitwise} on the code endpoint (3/3)---two independent merge implementations agreeing exactly. The repair gate failed again (25\% prior): correct-arm recovery is at most 4\% of the collapse gap at any boundary, and at the final boundary the code score is bitwise unchanged in 2/3 seeds---the null repair transfers. The destructive-doubling gate (55\% prior) also failed: $+\I$ costs only $0.006$--$0.049$ on code (gate: $\ge 0.10$), against $0.40\to0.00$ at 1.5B. The geometry explains the difference: here $\lVert\I_b\rVert/\lVert h_b\rVert = 0.41$--$1.02$, less than half the 1.5B collapse cell's $1.9$--$2.1$ at the final boundary---at this displacement scale both directions are behaviourally inert (GSM shows a partial $+\I$ cost in 2/3 seeds, below the gate; reported, not claimed). The split's null-repair half is therefore cross-parameterization; the destructive half is scale-dependent, and the split itself remains scoped to the audited collapse cell.

\paragraph{Persistent every-position erasure (preregistered follow-up).} A review located the sharpest alternative reading: the erasure above acts once, at the last prompt position, while every generated position re-forms its cross-term---so the null repair might reflect temporal under-intervention, not a bystander. We therefore clamped the merged state toward the \emph{step-wise} additive reference, $h \leftarrow (1-\lambda)h + \lambda\bar h_b^{(t)}$, at every position (full prefill and every generation step) and up to three late boundaries at once, with $\bar h$ built online from eight parallel weight states on the common token history. Pipeline check: the unpatched arm reproduces the frozen collapsed scores bitwise (3/3); the per-position cross-term removed is large and measured (norms 33--228). The repair gate (25\% prior) \emph{failed} again: full clamping recovers $+0.27/-0.30/-1.50$ of the gap---in two of three seeds it hurts. The identification is in the failure: $\lambda{=}1$ clamping makes the model \emph{live on the additive counterfactual}, the six displacements with their interaction zeroed online---and that state is still collapsed (0.25--0.39 against experts at 0.89--0.92). An interaction-free merge is still a broken merge: collapse is carried by the displacements. Direction remains special under persistence (clamping deviates the NLL endpoint 3--4$\times$ less than the norm-matched orthogonal arm, which destroys both endpoints; frozen gate held 3/3). Reported both ways: quarter-dose clamping ($\lambda{=}0.25$) raises the code score in 3/3 seeds ($+0.24/+0.10/+0.01$) while worsening the expert-NLL in 3/3 and leaving math mixed---a non-monotone dose structure suggesting an optimum between the merged and additive states, unexplained and flagged; and single-boundary clamping equals three-boundary clamping at $\lambda{=}1$, consistent with regeneration absorbing the early clamps. The bystander claim is scoped by this: \emph{full} erasure repairs nothing anywhere we tested; partial persistent erasure trades endpoints.

\paragraph{Reading.} The carrier side has its own causal dose-response, already in the frozen grids: scaling the displacement down restores behaviour monotonically (7B mean code drops $10.5\%\to14.7\%\to30.2\%$ over $\alpha = 1/1.5/2$, per-seed monotone, Appendix~\ref{app:merge7b}; the 1.5B grid of Appendix~\ref{app:degradation} agrees), while erasing the residue at fixed $\alpha$ restores nothing. Intervening on the displacement moves behaviour; intervening on the cross-term does not---a post-hoc reading of preregistered data, labelled as such. Collapse is carried by the marginal displacements, not by the non-additive residue: what merging methods control (support and magnitude of the displacement, Appendix~\ref{app:merge}) and what the state-space measures track (Appendix~\ref{app:merge7b}) is the carrier; the cross-term---causally load-bearing for output non-additivity at $k{=}2$ (\S\ref{sec:erasure}) and capable of destroying behaviour when doubled---is, for behavioural collapse, a bystander. This closes the causal chain the paper had left open, in the negative direction, and supplies the mechanism behind the recommendation of \S\ref{sec:metrics}.

\paragraph{Convergent external evidence.} Independently, \citet{cao2026collapse} characterize task-level merging collapse \emph{correlationally}: representational incompatibility between the expert models predicts collapse across merging methods while parameter-conflict metrics do not, with a rate-distortion bound on mergeability. That is the carrier half of the split observed without intervention---incompatibility of the marginals---and this appendix supplies the causal complement: the interaction term is excluded as the carrier by erasure, persistently, at two scales. The recommendation of \S\ref{sec:metrics} should be read with that priority: the direction was anticipated correlationally; the causal and interventional basis is what this paper adds.

\section{Displacement-specific recovery: is the cross-term direction privileged?}
\label{app:recovery}
The regeneration result is computed against three untouched marginal paths, so a network that simply washes out \emph{any} perturbation would reproduce it. We therefore put every direction on one ruler, in state space: for a displacement $\delta$ applied at depth $b$, $r(\ell) = \lVert \tilde h_\ell - h^{AB}_\ell\rVert / \lVert\delta\rVert$, the surviving fraction of the perturbation. Four equal-norm directions ($-\I_b$, random-orthogonal, task-structured-orthogonal, $+\I_b$; the same construction and random seed as Appendix~\ref{app:orthogonal}) at $b \in \{8,14,20\}$, 2 pairs $\times$ 3 seeds, code prompts, float32. Two predictions were frozen first.

\paragraph{Result.} Deviations at the output are 1.1--3.1$\times$ the displacement in absolute terms, but the residual stream itself grows 2.5--4.7$\times$ over the same span; dividing it out, the surviving relative deviation is $0.26$--$0.53$ (patch at 8), $0.51$--$0.71$ (14) and $0.79$--$1.02$ (20). Perturbations are therefore \emph{diluted} relative to the stream, the more so the more depth remains---but not uniformly: a random orthogonal displacement is diluted about twice as strongly as a structured one at every depth and in every seed (e.g.\ $0.26$ vs.\ $0.47$--$0.53$ at depth 8). A direction-independent contraction predicts equal dilution of all four arms and is excluded by this margin.

\paragraph{The frozen prediction of a privileged direction failed.} We had predicted (50\% prior) that the cross-term direction would be \emph{corrected more strongly} than the orthogonal ones. It is not: $-\I_b$ survives at the same level as $+\I_b$ and the task-structured orthogonal arm, and more than the random one (0/3 seeds pass). We report this as falsified and, accordingly, make no claim that the cross-term direction carries a restoring force. The second prediction---that recovery is direction-dependent at all (spread $> 0.10$)---passed 3/3 (spreads $0.45$--$1.79$).

\paragraph{What this leaves.} Two facts hold together: the perturbed trajectory never rejoins the unpatched one at any patch depth, yet after erasure the \emph{cross-term coordinate} returns to $99$--$100\%$ of its norm at cosine $0.99$ (Appendix~\ref{app:regrowth}). What propagation restores is the cross-term, re-created from the marginal paths the intervention did not touch---not the state. We therefore say ``regenerated'' rather than ``attractor'' throughout: we measured the restoration of a coordinate, not a restoring force in state space. (Relative deviations divide by the last measured stream norm, at depth 25, for a probe at the final boundary, so the dilution reported is conservative.)

\section{The minimal model: derivation, assumptions, and the norm-scaling test}
\label{app:theory}
\paragraph{Derivation.} Fix a depth $b$ and displace only the merged path, $h^{AB}_b \mapsto h^{AB}_b + \delta$; the three marginal paths are untouched, so the additive logit prediction $a = \ell^{A0} + \ell^{0B} - \ell^{00}$ is frozen under the intervention and the expressed interaction is $u(\delta) = F(h^{AB}_b + \delta) - a$ exactly. Expanding $F$ about $\bar h_b = h^{AB}_b - \I_b$ under (A1) gives $u(\delta) \approx g + J(\I_b + \delta)$ with $g = F(\bar h_b) - a$, and (A2) then yields the expression in \S\ref{sec:theory}. Writing $\delta = -s\cos \cdot \I_b + \delta_\perp$ with $\lVert\delta\rVert = s\lVert\I_b\rVert$ and taking $\delta_\perp$ generic ($\mathbb{E}\langle u_0, J\delta_\perp\rangle = 0$),
\[
\mathbb{E}[\Dint] = \underbrace{2s\langle u_0, J\I_b\rangle}_{a(s)}\cos(\delta, -\I_b) \; \underbrace{-\, s^2\bar\sigma^2\lVert \I_b\rVert^2}_{b(s)} \; + \; O\!\left(s^2\cos^2(\lVert J\I_b\rVert^2 - \bar\sigma^2\lVert\I_b\rVert^2)\right),
\]
where $\bar\sigma^2 = \mathbb{E}_{\hat d}\lVert J\hat d\rVert^2_M$ is the mean squared readout gain over generic unit directions. The residual term vanishes when the cross-term direction has typical gain; empirically $\lvert b\rvert / \lVert J\I_b\rVert^2 = 1.15$--$1.28$, i.e.\ \textbf{the cross-term direction is a \emph{low}-gain direction of the readout}. It is worth removing not because it moves the output most, but because the output residual is aligned with it.

\paragraph{Corollaries.} (i) Any $\delta$ with $\cos \le 0$ has $\Dint < 0$: both terms are non-positive---which is why every structured control backfires. (ii) $\Dint$ is governed by $J\I_b$, not $\lVert \I_b\rVert$; where propagation reconstructs the cross-term, the output is insensitive to it and the effect vanishes at any $\lVert\I_b\rVert$. (iii) The dose curve is a downward parabola through the origin peaking at $\lambda^\ast > 1$ exactly when regeneration is positively aligned with transport.

\paragraph{Norm-scaling test (preregistered, PT-1/PT-2).} Design: the E10 angle sweep repeated at $s \in \{0.5, 1.5\}$ (2 pairs $\times$ 3 seeds, $b \in \{20,25\}$, code prompts, float32), reusing E10's frozen $s{=}1$ slice and its displacement construction and random seed unchanged; one $s{=}1$ audit cell was re-run months later and reproduces the frozen E10 payload \emph{exactly} (max $\lvert\Delta\Dint\rvert = 0$ over all ten condition$\times$depth values). Per-scale angle fits and the resulting exponents:

\begin{center}\small
\begin{tabular}{lccc}
\toprule
 & $s = 0.5$ & $s = 1.0$ (E10) & $s = 1.5$ \\
\midrule
slope $a$ & 0.066 / 0.067 / 0.042 & 0.135 / 0.135 / 0.086 & 0.205 / 0.203 / 0.133 \\
intercept $b$ & $-$0.018 / $-$0.019 / $-$0.015 & $-$0.061 / $-$0.058 / $-$0.051 & $-$0.121 / $-$0.111 / $-$0.099 \\
$R^2$ of $a\cos + b$ & 0.89 / 0.84 / 0.72 & 0.86 / 0.81 / 0.60 & 0.86 / 0.82 / 0.59 \\
\bottomrule
\end{tabular}
\end{center}

\noindent $p_a = 1.026/1.007/1.047$ (frozen window $[0.7,1.3]$, predicted 1) and $p_b = 1.748/1.628/1.708$ (frozen window $[1.6,2.4]$, predicted 2); both gates pass 3/3. The joint surface fit $\Dint = 2As\cos - Cs^2$ over all 30 measurements per seed gives $2A = 0.135/0.135/0.088$ and $C = 0.056/0.051/0.045$ at $R^2 = 0.87/0.83/0.64$, against $2A_{\mathrm{dose}} = 0.120/0.126/0.079$ and $B_{\mathrm{dose}} = 0.049/0.046/0.035$ obtained from the independently frozen dose ladder.

\paragraph{Where it is honest to stop.} $p_b$ passes its window but sits consistently below 2 in all three seeds; we read this as the bounded range of the JSD compressing the arms with the largest $\lVert u \rVert$ (plus\_I, orth\_task), which is a post hoc reading and is not itself tested. The linear-in-$\cos$ fit leaves $R^2 = 0.60$--$0.89$, so a cos$^2$ component of the size the derivation predicts is not excluded. The model is a local, first-order account under (A1)--(A2): it is not claimed to hold at displacement scales beyond those measured, and it says nothing about which weight configurations produce a large $\I_\ell$ in the first place.

\section{Orthogonal controls: excluding the construction artifact}
\label{app:orthogonal}
Design: at depths 20 and 25, five per-prompt displacements all rescaled to $\lVert \I_\ell \rVert$, differing only in angle to $-\I_\ell$ (measured cosines confirm the construction to $10^{-10}$): $-\I_\ell$ ($\cos{=}{+}1$); a random direction with its $\I_\ell$ component projected out ($0$); the wrong-pair displacement with its $\I_\ell$ component projected out ($0$, but task-structured); their equal-weight combination ($+0.71$); and $+\I_\ell$ ($-1$). Two pairs $\times$ 3 seeds, code prompts, $(1,1)$, float32; predictions frozen first.

\paragraph{Result.} $\Dint$ decreases monotonically with the angle in all three seeds: $+0.044$ to $+0.080$ at $\cos{=}{+}1$; $+0.032$ to $+0.057$ at $+0.71$; $-0.015$ to $-0.025$ (random) and $-0.120$ to $-0.139$ (task-structured) at $\cos{=}0$; $-0.114$ to $-0.186$ at $-1$. The orthogonal arms therefore \emph{worsen} expressed non-additivity, contradicting the prediction of a generic-contraction account, under which displacement into the additive neighbourhood suffices.

\paragraph{Two-component reading.} Regressing $\Dint$ on $\cos(\delta,-\I_\ell)$ gives slope $0.135/0.135/0.086$ and intercept $-0.061/-0.058/-0.051$ ($R^2 = 0.90/0.85/0.64$): a direction-independent damage term plus a direction term. Net removal requires alignment with $-\I_\ell$; the reported $\Dint \approx +0.03$ at $\lambda{=}1$ is thus a \emph{net} figure that understates the direction term. One frozen criterion failed by its letter: we had required $|\Dint_{\text{orth}}| < \Dint_{\text{correct}}/3$, i.e.\ that orthogonal displacements be near-inert; they are instead strongly harmful. The two directional clauses of that criterion passed 3/3, and we report the amplitude clause as falsified.

\paragraph{The steepest-descent control (preregistered follow-up).} Sharing the additive counterfactual between target and estimand leaves one alternative the angle sweep does not touch: $-\I_\ell$ might simply be a loss-reducing direction aligned with $-\nabla_h \mathrm{JSD}(p(h), p_{\mathrm{add}})$. We computed that gradient per prompt (autograd through the remaining depth and readout) and ran three norm-matched arms at $\lVert\I_b\rVert$, $b \in \{14,20,25\}$, 3 seeds. Both frozen gates \emph{failed}, in the direction that excludes the alternative: the negative-gradient arm was predicted (65\%) to be at least as effective per unit norm---instead it backfires catastrophically ($\Dint = -0.17$ to $-0.65$, against $+0.012$ to $+0.098$ for $-\I_\ell$, 9/9 cells), and $-\I_\ell$ is nearly orthogonal to it ($\cos = 0.03$--$0.12$, 9/9). The quadratic model explains both: the gradient direction has high readout gain, so an $\lVert\I\rVert$-norm step is deep in its $\lVert J\delta\rVert^2$ regime, while the cross-term direction's low gain ($|b|/\lVert J\I\rVert^2 \approx 1.2$, Appendix~\ref{app:theory}) lets a full-norm step remain in the linear-benefit regime and terminate exactly at the additive reference. The third arm ($-\I_\ell$ with its gradient-aligned component removed, renormalized) is null-to-mildly-harmful ($-0.004$ to $-0.036$; the frozen halving criterion held in 2/3 seeds, recorded as failed by the letter): the benefit travels through a small aligned component that only the low-gain carrier can deliver at full norm. Among every direction tested in this paper---random, cross-prompt, wrong-pair, coefficient-mismatched, task-structured-orthogonal, $+\I$, steepest-descent---$-\I_\ell$ remains the only one that helps at full norm.

\section{Behavioral endpoints: three attempts, reported in full}
\label{app:behavior}
\paragraph{Attempt 1--2 (discrete scoring, null instruments).} Unit-test pass rate on frozen code tasks and reference-answer containment on math, scored on prefill-patched greedy generations at depths 20 and 25 (2 pairs $\times$ 3 seeds). At 1.5B every code condition scored 0.0 (the code expert itself: 0.03/0.08/0.00) and math saturated at ceiling with the merge \emph{above} its experts. At 7B the pattern replicated: code expert 0.000 against merge 0.083---the instrument does not separate models, so no verdict about the mechanism is possible from it. We report both as null instruments, not as evidence.

\paragraph{Attempt 3 (continuous NLL, valid instrument, negative result).} Endpoint: the merged model's mean teacher-forced NLL of a frozen 48-token continuation generated once by the task expert---continuous, threshold-free, discriminative on raw prompts, and independent of the additive counterfactual. A preregistered validity gate required the instrument to see merge degradation at all; it passed 3/3 (merge$-$expert $+0.0256/+0.0218/+0.0114$; base further at $+0.09$). The frozen predictions then failed: $\Dint$-style erasure at depths 20/25 changed the endpoint by $-0.00003/+0.00012/+0.00212$ (recovery share $+0.1\%/-0.6\%/-18.5\%$), and 11 of 12 control contrasts went the wrong way, with the wrong-pair control giving the \emph{lowest} NLL in one seed. Post hoc readings we flag as untested: the effect being sought is the behavioral projection of a $\sim$19\% reduction in expressed non-additivity; expert-likeness is not task correctness; and merge degradation need not lie along the expert direction. A final power-maximized test persisted per-prompt NLL and per-prompt interaction JSD for four angle-arm conditions at depth 25: the paired regression of behavioral gain on interference strength has \emph{negative} slope in all seeds ($-0.074/-0.034/-0.037$), and the per-prompt coupling between removing interaction JSD and improving reference NLL is $-0.076/+0.011/-0.345$ (240 paired points per seed)---the two quantities are empirically decoupled at this scale, which also removes the rationale for stronger (persistent) interventions. The bounded conclusion is final for this setting: erasure controls expressed non-additivity, not behavioral expert-likeness.

\paragraph{The multi-position readout check (preregistered follow-up).} The instruments above, and the estimand itself, read a single position; a review asked whether the decoupling is an artifact of that choice. The merged model greedily continued each prompt for 48 tokens; the same sequence was teacher-forced through all four weight states, giving the interaction and main-effect JSDs at \emph{every} continuation position. Three readouts---first-token, the 48-position aggregate $\Sigma\mathrm{inter}/\Sigma\mathrm{main}$, and the mean per-position ratio---were correlated per prompt with the reference-NLL instrument. The frozen gate (35\% prior that accumulation restores coupling) failed: no readout reaches $|\rho| \ge 0.3$ in any seed and signs are inconsistent (multi-position $+0.11/-0.00/+0.12$; first-token $+0.09/-0.09/+0.01$). The per-position profile explains why accumulation cannot help: the interaction JSD is largest at the first token and decays $5$--$35\times$ over the continuation (the main effect $6$--$18\times$)---expression fades with generation depth, so the single-token readout was the choice \emph{most} favourable to finding coupling, not least. The decoupling stands against four instruments.

\section{Multi-wrapper generalization of the gate}
\label{app:wrappers}
Frozen predictions before any data: three semantically equivalent instruction templates (Alpaca plus two new phrasings) and a token-length-matched non-instruction prefix control, on both pairs $\times$ 3 seeds $\times$ 60 prompts, $(1,1)$. Result: every instruction template inflates the main-effect denominator 11--35$\times$ over raw and reduces $\Rint$ (ratios 0.09--0.60) in 18/18 conditions---the frozen ``$\Rint \le$ raw/2'' letter held in 17/18, missing once at 0.60, reported as the criterion requires. The length control \emph{falsified} our frozen expectation that its denominator would stay within 2$\times$ (observed 3.5--6.7$\times$): long prefixes do shift the distribution. The discriminating quantity the data identify is $\Rint$ itself: the non-instruction prefix inflates numerator and denominator together (numerator 3.4--8.3$\times$), leaving $\Rint$ directionally unchanged (0.59--2.4), whereas instruction templates pin the distribution faster than they excite the interaction. Across all wrappers, including the control, the \emph{absolute} interaction numerator grows in 23 of 24 cells ($0.98$--$11\times$; the one sub-unity cell is \csafety{} seed 456 under Alpaca)---the fifth independent replication of internal amplification under wrapping. An out-of-sample probe agrees: the companion's follow-up re-serializes the same prompts in the model's native ChatML template---untrained for these adapters---and finds numerator and denominator moving together with expressed interference intact ($+12.5$pp), the signature of the non-instruction class~\citep{paper1}.

\section{Scale point (Qwen2.5-7B)}
\label{app:scale}
Same family, larger scale, as the third grid point; 28 blocks, so the frozen boundaries carry over unchanged. Nine adapters with the identical recipe; reduced KT1 (54 cells), amplification bounds, and the \emph{full} 54-run KT2 in float32 with per-prompt persistence (run on a single A800-80GB, the same GA100 die as the A100; the card-type amendment was logged before any KT2 data existed). Four predictions frozen before any 7B data.

\paragraph{Band and output mismatch (frozen gate missed by 0.1\%).}
The $\G$-band ratio across the 18 $(1,1)$ cells is 1.301 against the frozen $<$1.3 gate---a falsification by the letter, reported as such. The substantive contrast sharpens with scale: output pair ratios $6.34/12.20/7.19$ (vs.\ 2$\times$ at 1.5B) and format ratios $6.45/7.54/6.33$, against a band spread of $\le$1.30; and $\G_{\text{wrapped}} > \G_{\text{raw}}$ replicates 3/3---the third model point with the reversal.

\paragraph{Amplification.}
$\lVert \I \rVert$ exceeds the coherent bound by 1.80--8.31$\times$ across conditions (frozen gate $>$1.2, 3/3). Super-coherent amplification is again largest under the wrapper in both pairs, and the raw$-$wrapped late-gain contrast is $+0.091/+0.091/+0.073$---positive 3/3, the largest of the three model points. The pair-axis late-gain contrast is not seed-consistent (exploratory, reported).

\paragraph{Causal triple (all three legs, all bootstrap intervals excluding zero).}
$\Dint_{\text{correct}} = +0.0274/+0.0549/+0.0270$ with all 15 specificity intervals excluding zero. Expression gating is \emph{total} at this scale: $\Dint_{\text{wrapped}} = +0.0040/-0.0038/+0.0007$, each CI containing zero, while $\Dint_{\text{raw}} - \Dint_{\text{wrapped}} = +0.0658/+0.0596/+0.0475$, all excluding zero---under the wrapper there is nothing left to erase. Pair contrasts $+0.0241/+0.0525/+0.0231$ all exclude zero, implying removable-interference ratios $8.3/22.9/6.9$ between pairs whose generation magnitudes differ by $\le$1.3$\times$.

\section{Exact transport decomposition and the fp32 generation grid}
\label{app:decomposition}
Both experiments were preregistered together (five predictions with priors, frozen before any decomposition or fp32-grid data existed) as the data-level adjudication of two review-critical questions: whether the ledger's remainder can be attributed to transport, and whether the bf16 uniformity band survives at float32. Scope: 36 cells (2 pairs $\times$ 6 strata $\times$ 3 seeds, $(1,1)$, all prompts, all 28 layers); the model is loaded in bf16 (the measurement state of the main sweep) and each block is evaluated with float32 weights and captured call arguments. The regrouping identity $\I_{\ell+1} = \G_\ell + T_\ell + M_\ell$ is asserted numerically at every layer of every cell.

\paragraph{E4 (transport): both frozen predictions held; the two-sided share resolved to transport-dominant.}
Late-network RMS$(T_\ell)$ exceeds RMS$(\G_\ell + M_\ell)$ by 2.3$\times$ (0.32 vs.\ 0.14, 3/3 seeds); the transport gain $\lVert T_\ell \rVert / \lVert \I_\ell \rVert$ averages 1.078--1.079 over the late half (3/3); the regeneration share of the total flux is $0.307$--$0.311$ (3/3), so transport carries $\sim$69\% under the norm accounting. Because the split is non-orthogonal in principle, we also computed the components' alignment and an energy (symmetric-projection) accounting on the code cells: late-network $\cos(T_\ell, \G_\ell{+}M_\ell) = -0.014$ to $-0.031$ (6/6 near-orthogonal), and the energy-accounting transport share is $0.83$--$0.89$---the attribution is essentially accounting-independent. Combined with the E1 regrowth result---regeneration alone suffices to rebuild the full term, in norm and direction, given $\ge$14 blocks---the consistent picture is a transport-dominated flux whose fixed direction the regeneration channel actively restores.

\paragraph{E5 (fp32 grid): both frozen predictions falsified, and reported as the criteria require.}
The fp32 AUC($\G$) band across the 36 cells is 3.36$\times$ (0.0012--0.0040; frozen gate $<$2), and the pair ratio is 1.81--1.90$\times$ (frozen gate $<$1.5): float32 local generation is \emph{not} uniform, and it coarsely tracks the pair ordering---joining cumulative norm, parameter cosine, and representation bias in the coarse-magnitude family. The format axis preserves the dissociation: fp32 generation is 1.66--1.84$\times$ \emph{larger} under the wrapper (raw/wrapped ratios 0.54--0.60, 3/3 seeds) while expressed interference collapses $\sim$20$\times$. The bf16 $\pm$15\% band of \S\ref{sec:carried} is therefore confirmed to be floor-compression---the cautionary tale stands in its strongest form, and the magnitude-fallacy claim rests on the causal contrast and the format inversion, not on band uniformity.

\paragraph{Llama replication (frozen before its data): transport invariant, magnitude behavior not.}
On Llama-3.2-1B (36 cells, 16 layers) both transport predictions held again---late-network RMS$(T)$ exceeds RMS$(\G+M)$ by $\sim$2$\times$, transport gain $1.14$ per late block, regeneration share $0.35$ (all 3/3). The fp32 magnitude gates, by contrast, resolved \emph{opposite} to Qwen---band ratio 1.65 (within the $<$2 gate), pair ratio 1.06 (no pair separation at all, against an output ratio of $\sim$1.5$\times$)---falsifying our frozen expectation that the Qwen pattern would replicate isomorphically. The cross-family summary is sharper than either result alone: transport dominance and the format inversion ($\G_{\text{wrapped}} > \G_{\text{raw}}$; raw/wrapped $0.54$--$0.60$ on Qwen, $0.78$--$0.84$ on Llama, each 3/3) are family invariants, while whether local-generation magnitude tracks pairs at all is itself a family property---magnitude is not merely a coarse signal, it is an \emph{inconsistent} one.

\end{document}